\documentclass[10pt,twocolumn,letterpaper]{article}

\usepackage[publish]{cvpr} % To force page numbers, e.g. for an arXiv version
\usepackage{float}

\definecolor{cvprblue}{rgb}{0.21,0.49,0.74}
\usepackage[pagebackref,breaklinks,colorlinks,allcolors=cvprblue]{hyperref}
\usepackage{float}
\usepackage{multirow}
\usepackage{multicol}
\usepackage{makecell}
\usepackage[pagebackref,breaklinks,colorlinks,allcolors=cvprblue]{hyperref}
\usepackage{diagbox}
\usepackage{float}
\usepackage{multirow}
\usepackage{diagbox}
\usepackage{multirow}
\usepackage{booktabs}
\usepackage{graphicx}  % 用于\resizebox
\usepackage{bm} 
\def\paperID{} % *** Enter the Paper ID here
\def\confName{CVPR}
\def\confYear{2026}

\newcommand{\tyz}[1]{\textcolor{black}{#1}}
\newcommand{\lqr}[1]{\textcolor{black}{#1}}
\newcommand{\yr}{\textcolor{black}}
\newcommand{\best}[1]{\textcolor{red}{#1}}
\newcommand{\second}[1]{\textcolor{blue}{#1}}

\title{InpaintDPO: Mitigating Spatial Relationship Hallucinations in Foreground-conditioned Inpainting via Diverse Preference Optimization}

\author{
Qirui Li$^{1}$\footnotemark[1] ~~
Yizhe Tang$^{1}$\footnotemark[1] ~~
Ran Yi$^{1}$\footnotemark[2] ~~
Guangben Lu$^{2}$ \\
Fangyuan Zou$^{2}$ ~~
Peng Shu$^{2}$ ~~
Huan Yu$^{2}$ ~~
Jie Jiang$^{2}$ \\ 
$^1$Shanghai Jiao Tong University \quad $^2$Tencent
}

\begin{document}

\twocolumn[{%
\maketitle
\begin{figure}[H]
\vspace{-0.4in}
\hsize=\textwidth 
\centering
\includegraphics[width=0.87\textwidth]{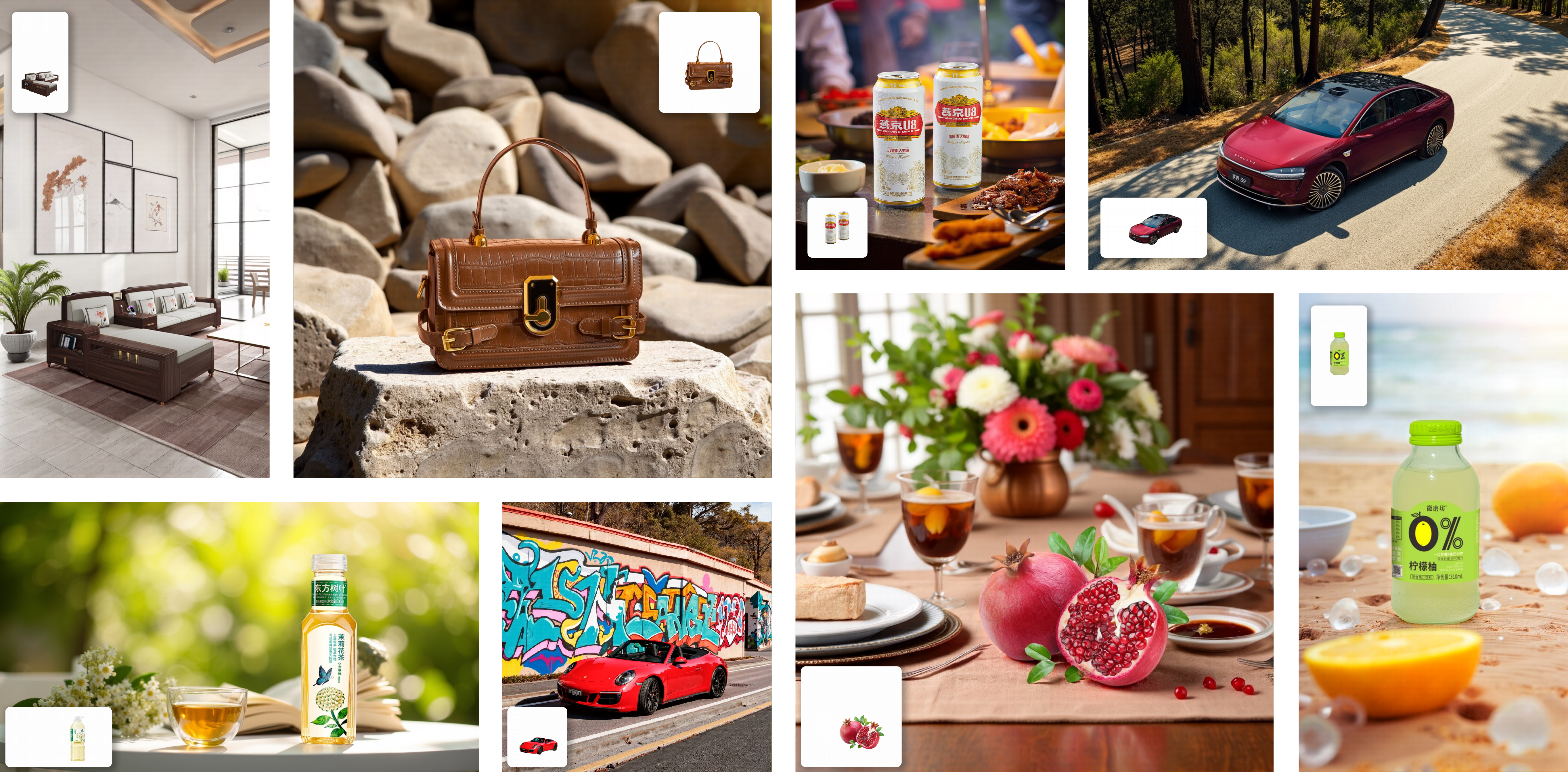}
% \vspace{-0.2in}
%\caption{\yrn{Generated} results of our Pinco Model. Pinco \yrn{achieves} high-quality \yrn{generation of rich and diverse} backgrounds.}
\caption{Our \textbf{InpaintDPO} generates high-quality images with plausible spatial relationships between foreground subject and background scene, \yr{while also achieving} outstanding foreground consistency and text alignment.}
% \vspace{-0.1in}
\end{figure}
}]

{
  \renewcommand{\thefootnote}%
    {\fnsymbol{footnote}}
  \footnotetext[1]{Equal contribution.} %\quad $\{$lucasgblu, yuzhendu$\}$@tencent.com}
  \footnotetext[2]{Corresponding author. \quad ranyi@sjtu.edu.cn}
}

\begin{abstract}
Foreground-conditioned inpainting, which aims at generating a harmonious background for a given foreground subject based on the text prompt, is an important subfield in controllable image generation.
A common challenge in current methods, however, is the occurrence of Spatial Relationship Hallucinations between the foreground subject and the generated background, including inappropriate scale, positional relationships, and viewpoints. 
Critically, the subjective nature of spatial rationality makes it challenging to quantify, hindering the use of traditional reward-based RLHF methods.
To address this issue, we propose InpaintDPO, 
the first Direct Preference Optimization (DPO) based framework dedicated to
spatial rationality in foreground-conditioned inpainting, ensuring plausible spatial relationships between foreground and background elements. 
To resolve the gradient conflicts in standard DPO caused by identical foreground in win-lose pairs, 
we propose MaskDPO, which confines preference optimization exclusively to the background to enhance background spatial relationships, while retaining the inpainting loss in the foreground region for robust foreground preservation. 
To enhance coherence at the foreground-background boundary,
we propose Conditional Asymmetric Preference Optimization, which samples pairs with differentiated cropping operations and applies global preference optimization to promote contextual awareness and enhance boundary coherence.
Finally, based on the observation that winning samples share a commonality in plausible spatial relationships, we propose Shared Commonality Preference Optimization to enhance the model’s understanding of spatial commonality across high-quality winning samples, further promoting shared spatial rationality.
Extensive experiments demonstrate that InpaintDPO significantly mitigates Spatial Relationship Hallucinations, achieving superior performance in background spatial rationality while ensuring competitive foreground preservation and text-alignment.
The code and models of our migration on Qwen-Image-Edit will soon be available at \url{https://github.com/IApple233/InpaintDPO}.
\end{abstract}
\section{Introduction}
\label{sec:intro}
Controllable image generation~\cite{controlnet,tan2025ominicontrol,hu2025high,ye2023ip,yi2024feditnet,hu2025improving,yi2025iar2,bu2025hiflow} has seen remarkable progress, enabling the creation of highly realistic and diverse images. 
A crucial sub-field, %image inpainting, has also made significant progress, particularly in filling or replacing image regions under textual prompt. 
foreground-conditioned background inpainting~\cite{lu2025pinco,tang2025ata,chen2025ctr} stands out as a specialized %yet highly practical task,
\yr{task with direct applications in areas such as advertisement design and product promotion. It}
focuses on generating a text-aligned and contextually appropriate background for a given foreground object. 
%Its utility is especially pronounced in commercial applications such as product background synthesis for e-commerce.

\begin{figure}[t]
    \centering
    \vspace{-0.1in}
    \includegraphics[width=0.9\linewidth]{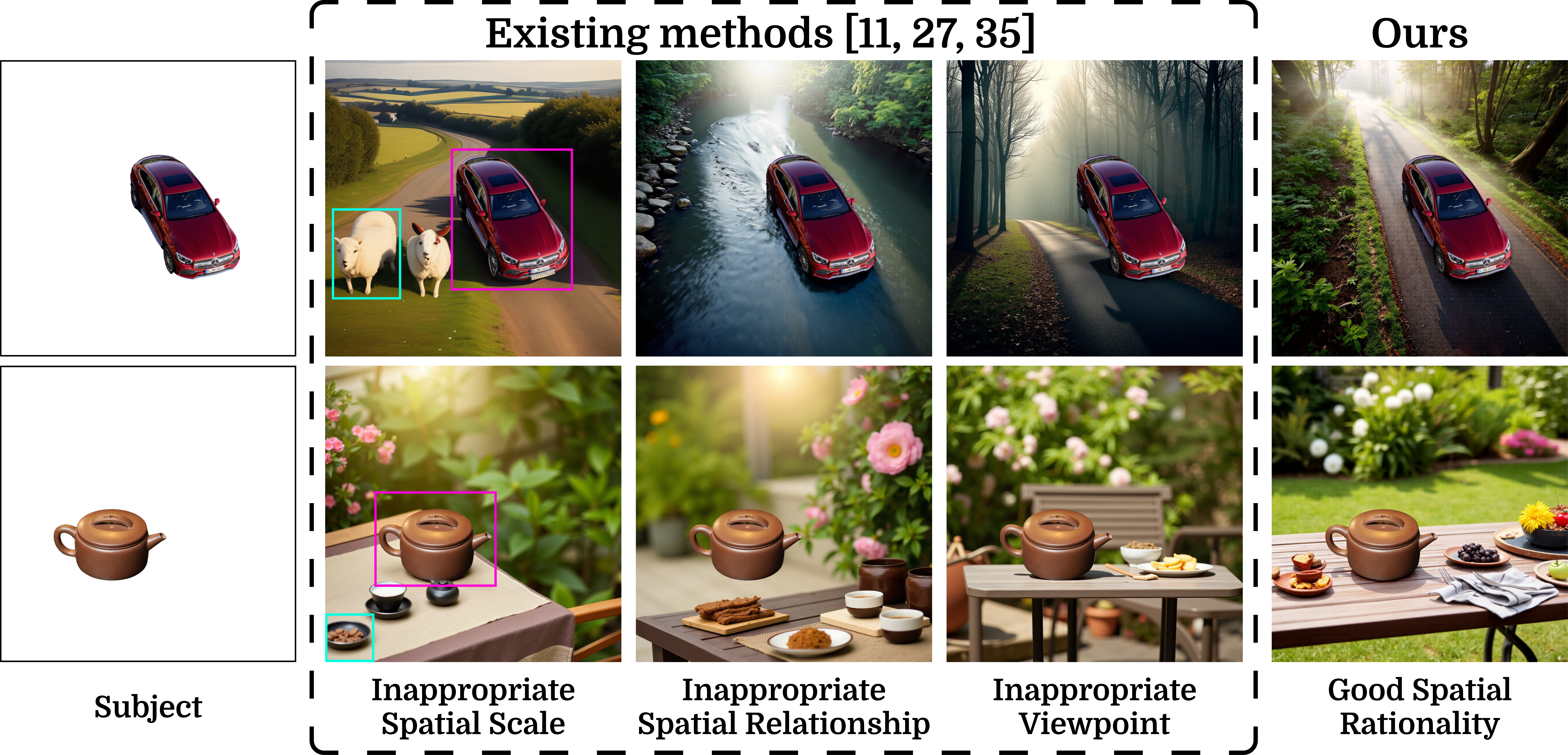}
    % \vspace{-0.1in}
    \caption{Three categories of ``\textbf{Spatial Relationship Hallucinations}" in foreground-conditioned inpainting, where previous methods~\cite{flux2024inpaint,alimama_FLUX,lu2025pinco} suffer from inappropriate spatial scale% (the size of foreground is irrational relative to background element)
    , spatial relationship % (the foreground subject is floating on water or in air)
    and viewpoint. % (distinct viewpoint of foreground and background)
    \yr{In contrast,} our method can generate spatially rational images without such hallucinations.}
    \vspace{-0.2in}
    \label{fig: hallucination}
\end{figure}

%Despite this progress, a prevalent issue in previous foreground-conditioned inpainting methods 
\yr{However, a prevalent issue in existing foreground-conditioned inpainting methods}
is the overemphasis on preserving the foreground features, which disrupts the model's originally learned distribution, leading to “\textbf{Spatial Relationship Hallucinations}'' where the generated background elements are positioned illogically relative to the foreground as in Fig.~\ref{fig: hallucination}. 
These hallucinations can be classified into three categories: 1) Inappropriate Spatial Scale, where the foreground object often exhibit\yr{s} incorrect relative sizes compared to background references; 
2) Inappropriate Positional Relationship, which is often observed as unsupported placement (objects appearing to float without physical support) or geometric errors (contact surface misalignment);
3) Inappropriate Viewpoint, where the camera viewpoint of the foreground mismatches that of the background scene.

Recently, with the \yr{successful} application of RLHF in LLM~\cite{achiam2023gpt,ouyang2022training,christiano2017deep}, some T2I methods~\cite{wallace2024diffusion,black2023training,fan2023dpok} have also attempted to use RLHF strategies to generate results that are more aligned with human preferences, especially based on a pretrained Reward Model~\cite{zhang2019bertscore,xu2023imagereward,radford2021learning} to improve overall aesthetic appeal or objective aspects, such as counting accuracy and prompt-image alignment.
However, in our task, the core challenge is that the \textit{spatial rationality} of an image is inherently subjective and difficult to quantify with any metric or pretrained model. 
This difficulty hinders the use of reward-based reinforcement learning strategies like PPO~\cite{schulman2017proximal,black2023training} or GRPO~\cite{shao2024deepseekmath,xue2025dancegrpo}, which require a clear, computable reward signal. 
As a result, we turn to \yr{design} a novel post-training methodology \yr{for spatial rationality enhancement} based on Direct Preference Optimization (DPO)~\cite{rafailov2023direct}. 
We first construct a customized \yr{human} preference dataset by generating and filtering a large pool of background inpainting samples across diverse foreground subjects. 
Then we conduct rigorous manual annotation to establish a high-quality dataset of win-lose pairs based on the rationality of spatial relationships.
By fine-tuning the model using our DPO-based approach, we aim to directly align the generation process with human preferences, yielding background synthesis that is %logically sound and visually harmonious.
\tyz{spatially rational and visually harmonious.}

While successful in some Text-to-Image (T2I) domains~\cite{wallace2024diffusion,yang2024using,liang2025aesthetic}, directly applying standard DPO \yr{to enhance spatial rationality} in foreground-conditioned inpainting
% faces significant challenges 
\yr{faces a fundamental conflict.}
%The issue arises from the \yr{the special nature of the task: 
\yr{Since the task requires the foreground subject to be preserved as the given one, } %construction of the win-lose pairs, where 
the foreground remains identical across \yr{winning and losing samples in a} pair.
%When applying standard DPO, the identical foreground regions \yr{in both winning and losing samples} provide conflicting optimization gradients, 
\yr{Applying standard DPO, which optimizes the entire image, forces the model to simultaneously encourage and discourage the same foreground features. This conflicting signal}
results in model degradation %over training epochs, 
such as missing details or incomplete denoising in the subject area.

To address this challenge, we propose \textbf{InpaintDPO}, a novel \yr{post-training} method that %, for the first time, introduces human preference alignment based on Direct Preference Optimization (DPO) \yr{to} specifically \tyz{emphasize spatial rationality} for the task of foreground-conditioned inpainting.
\yr{establishes the first DPO-based framework dedicated to spatial rationality in foreground-conditioned inpainting, ensuring plausible spatial relationships between foreground and background elements.}

Firstly, to resolve the gradient conflict \yr{in standard DPO caused by} \yr{identical} foregrounds across the win-lose pairs, we propose \textbf{MaskDPO}, which leverages the foreground mask to %bypass 
\yr{confine} the DPO optimization %in the subject region, 
\yr{to the background,}
%ensuring that preference learning is applied exclusively to the background area. 
\yr{applying preference learning only where needed.}
Simultaneously, to guarantee robust foreground subject preservation, we \yr{retain} the original inpainting loss within the \yr{foreground} subject area. 
This \yr{spatially disentangled optimization} mechanism \yr{enables} MaskDPO \yr{to} refine background spatial relationships without compromising the preservation of the foreground subject. Secondly, \yr{to enhance coherence at the foreground-background boundary,}
we propose \textbf{Conditional Asymmetric Preference Optimization (CAPO)}, which %samples win-lose pairs by introducing \yr{a} differentiated cropping operation to the foreground subject \tyz{and then \yr{applies} a global preference optimization}.
\yr{performs global preference optimization to win-lose pairs that differ in foreground positioning, achieved through differentiated cropping.}
\tyz{This %sampling strategy ensures the foregrounds differ in their positions across win-lose pairs, 
%which prevents the gradient conflicts arising from global preference optimization, 
\yr{positional difference naturally prevents gradient conflict seen in standard DPO,}
while} \yr{the subsequent global preference optimization}
%a global preference optimization loss can be applied across the entire image, which  
effectively %enhancing 
\yr{enhances} the visual coherence at the foreground boundary. Thirdly, \tyz{we observe that winning samples %inherently share a commonality in plausible spatial relationship.
\yr{share common spatial relationship characteristics.}} 
%\tyz{Therefore,} to further promote the model's capacity to learn such commonality of spatial rationality from the win-sample pairs, 
\yr{To capture this commonality,}
we propose %\textbf{Preference Similarity Learning (PSL)}. 
\textbf{Shared Commonality Preference Optimization (SCPO)}.
%which aims to minimize the preference distance between two winning samples.
\yr{Its goal is to make the model's implicit rewards for high-quality winning samples not just high, but also concentrated.} \yr{By narrowing the implicit reward gap between win-win pairs,
\tyz{SCPO} explicitly teaches the model to capture and enhance the shared spatial relationship characteristics that lead to success.}

% This strategy employs a win-win loss concurrently with the DPO loss: while MaskDPO maximizes the preference gap between win-lose pairs, win-win Loss minimizes the prefence gap between win-win pairs. 
% This approach effectively captures shared spatial information among high-quality samples, significantly enhancing the model's spatial understanding and positional rationality.

Extensive comparative experiments %validate the effectiveness of our InpaintDPO approach, which not only demonstrates superior inpainting capabilities in foreground consistency and text-alignment, but also brings better performance on background spatial relationship rationality.
\yr{demonstrate that our InpaintDPO achieves significant gains in background spatial relationship rationality, while also maintaining superior performance in foreground consistency and text alignment.}

We summarize our contributions as follows:
\begin{enumerate}
    \item %To address \yr{the} spatial relationship hallucination problem in existing foreground-conditioned inpainting methods, 
    We propose \yr{InpaintDPO}, %a novel InpaintDPO framework 
    \yr{the first DPO-based framework to address spatial relationship hallucination in foreground-conditioned inpainting, and a dedicated}
    %introduce a specialized 
    human preference dataset for spatial rationality \yr{alignment}.
    \item We propose MaskDPO to \yr{resolve} \yr{standard} DPO's gradient conflict by %leveraging foreground mask to ensure the preference optimization \yr{is} only applied in background region 
    \yr{confining preference optimization only to background region, enhancing background spatial relationships}
    while preserving the foreground through inpainting loss.
    \item We propose Conditional Asymmetric Preference Optimization to enhance %visual 
    coherence at the foreground boundary 
    %by introducing a differentiated cropping operation to the foreground in sampling, 
    \yr{%It enables 
    \tyz{via} conflict-free global preference optimization with differentiated cropping,}
    %which prevents conflicts and 
    \yr{effectively promoting contextual awareness and enhancing boundary coherence.} %improving positional rationality.
    \item We propose %Preference Similarity Learning, 
    \yr{Shared Commonality Preference Optimization,}
    %a novel strategy 
    \yr{which captures} %to implicitly learn 
    the commonality of plausible spatial relationships %and promote shared spatial rationality 
    by narrowing %preference distance 
    \yr{implicit reward gap among} high-quality winning samples\yr{, further promoting} shared spatial rationality.
\end{enumerate}
\section{Related Work}
\label{sec:related}

\subsection{Image Inpainting}

With remarkable advancements in Text-to-Image (T2I) diffusion model~\cite{ldm,sdxl,dit,sd3,flux2024}, a variety of diffusion-based inpainting methods have been proposed~\cite{avrahami2023blended,ju2024brushnet,zhuang2023powerpaint,chen2023pixart,manukyan2023hd,eshratifar2024salient,lu2025pinco,hu2024anomalydiffusion}.
Current methods can be broadly categorized into three groups:
1) Sampling-based methods~\cite{avrahami2022blended, avrahami2023blended, chen2023pixart, corneanu2024latentpaint, wang2025towards} modify latent features during the diffusion process to incorporate given regions. 
But it often disrupts the original latent space distribution, leading to semantic inconsistencies or spatial artifacts.
2) Fine-tuning based methods~\cite{manukyan2023hd,zhuang2023powerpaint,tan2025ominicontrol,flux2024inpaint,yu2025omnipaint} adapt the pre-trained T2I model structure for specific inpainting tasks.
However, they are prone to catastrophic forgetting, which affects the pre-learned general knowledge, limiting the quality and diversity of generated images.
3) Condition-injecting methods~\cite{alimama_FLUX,he2024freeedit,ju2024brushnet,eshratifar2024salient,tang2025ata,chen2025ctr} inject features of given area into a frozen T2I backbone via an external branch. 
However, with the increasing scale of base models, the branch networks become larger, leading to parameter overhead and high computational costs.
Some methods~\cite{lu2025pinco,tang2025ata,zhang2025ultra} turn to train a light-weight adapter for effective condition injection.
But due to the lack of spatial information perception, sometimes their results contain spatial relationship hallucinations.
We base on the state-of-the-art foreground-conditioned inpainting model FLUX-Pinco~\cite{lu2025pinco} and perform a DPO-based post-training to achieve better spatial relationships.

\subsection{RLHF for T2I model}
Reinforcement Learning from Human Feedback (RLHF) is a paradigm designed to align model with complex human preferences~\cite{christiano2017deep,achiam2023gpt,kaufmann2024survey}, which mostly involves a Reward Model (RM) trained on human-annotated data and fine-tunes the generative model using the RM's predicted score as a learning signal.
With the widespread application of RLHF on LLM~\cite{rafailov2023direct,schulman2017proximal,shao2024deepseekmath,ouyang2022training}, some reinforcement learning based methods designed for T2I diffusion models have emerged~\cite{xu2023imagereward,wallace2024diffusion,black2023training,yang2024using,fan2023dpok}.
Current T2I RLHF methods can be categorized into two types based on their optimization strategy:
1) Reward-Model based methods~\cite{xu2023imagereward,black2023training,fan2023dpok,zhangitercomp,hu2025towards,xue2025dancegrpo,liu2025flow} usually pretrained a separate reward model~\cite{xu2023imagereward,zhang2019bertscore,radford2021learning,wu2023human,kirstain2023pick} based on human preference data or utilize Vision-Language Model to assign a scalar reward to generated images.
The model is then fine-tuned using RL techniques like PPO~\cite{schulman2017proximal} or GRPO~\cite{shao2024deepseekmath} to maximize this predicted reward signal.
While effective in aligning models to subjective human standards, this approach is limited by the accuracy and convergence of the reward model~\cite{clark2023directly}.
2) DPO-based methods~\cite{wallace2024diffusion,yang2024using,liang2025aesthetic,karthik2025scalable,huang2025patchdpo,li2024aligning} directly optimize the T2I model using human preference data by reframing the optimization as a maximum likelihood problem~\cite{rafailov2023direct} over the preference pairs, avoiding the explicit training and sampling steps required by the separate RM and RL phase.
This strategy is more computationally efficient and enables more stable training, but \yr{requires an} extensive collection of high-quality comparative human preference data.
\begin{figure*}[t]
    \centering
    \vspace{-0.2in}
\includegraphics[width=0.9\textwidth]{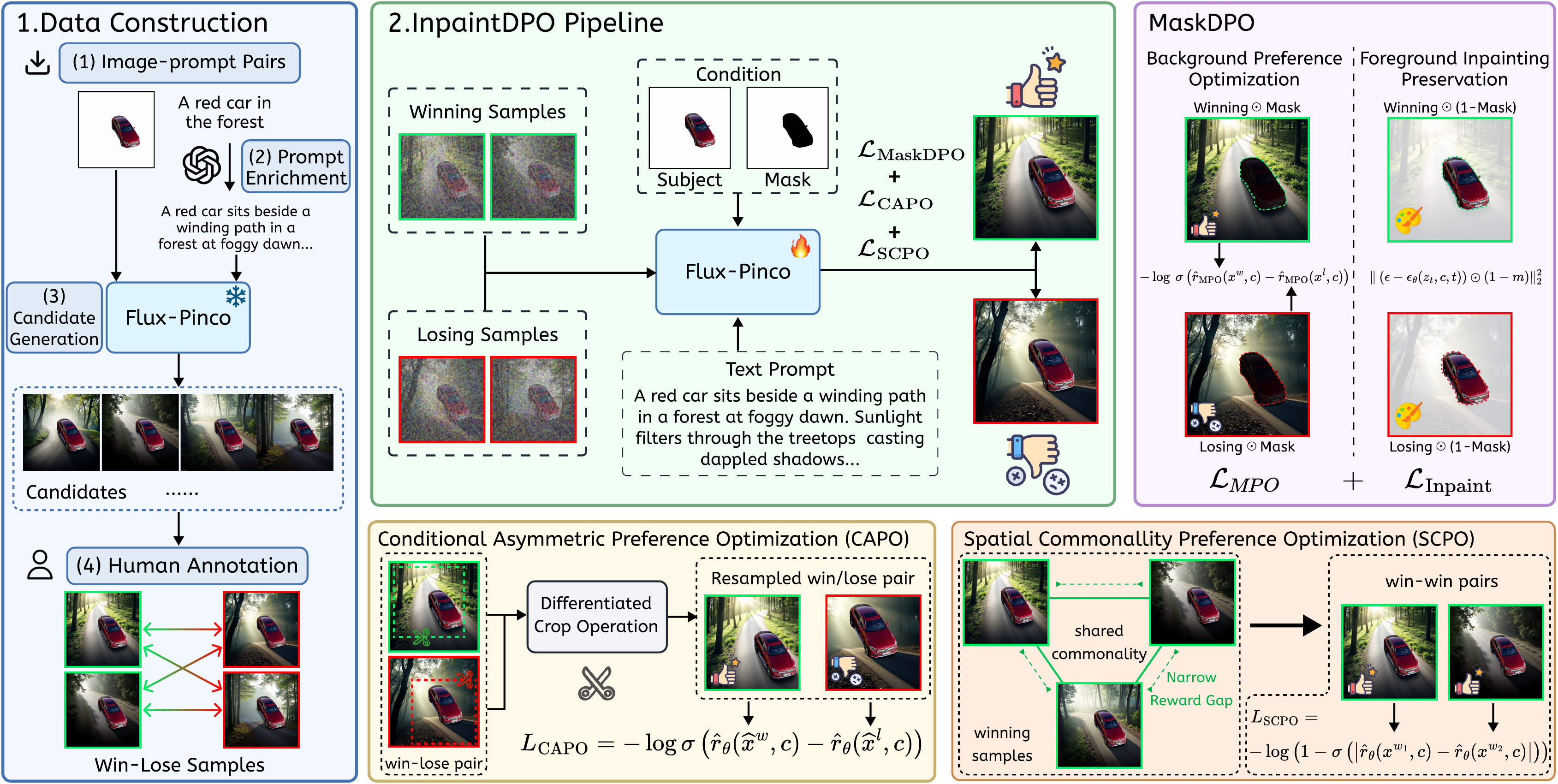} 
    % \vspace{-0.2in}
    \caption{Overview of \textbf{InpaintDPO} framework for foreground-conditioned inpainting.
    1) Data Construction: a four-stage pipeline to build the high-contrast spatial rationality human preference dataset.
    2) InpaintDPO training pipeline and three proposed preference optimization strategies: 
    MaskDPO confines preference optimization to background, while \yr{retaining} inpainting loss for foreground preservation;
    CAPO \yr{enhances} foreground boundary coherence through global preference optimization via differentiated cropping operation;
    SCPO captures the shared commonality of plausible spatial \yr{relationships} of winning sample\yr{s} by narrowing implicit reward gap.}
    \label{fig:model_archi}
    \vspace{-0.2in}
\end{figure*}

\section{Inpaint DPO}
\yr{Foreground-conditioned inpainting aims at generating a high-fidelity background conditioned on a text description, while preserving the integrity of the given foreground subject.} A primary challenge in this task is to ensure spatial rationality between the foreground and generated background.
Previous methods suffer from \textbf{Spatial Relationship Hallucinations}, generating illogical background elements manifesting as inappropriate scale, unsupported placement, and mismatched viewpoints.
Therefore, we turn to DPO, a human preference alignment technique \yr{that} directly optimizes on the preference likelihood without learning an explicit reward model. \tyz{However, directly applying standard DPO to our task faces a critical issue, where \yr{the identical foreground} across win\yr{-}lose pairs exhibit conflicting optimization gradients, leading to model degradation over training.}
To address \yr{this issue}, we propose \textbf{InpaintDPO}, a novel DPO-based approach specifically designed to promote realistic and harmonious spatial relationships between the foreground and generated background.
\tyz{Crucially, InpaintDPO is a flexible preference optimization framework applicable to any existing diffusion-based foreground-conditioned inpainting model.}

Specifically, to address the scarcity of foreground-conditioned win-lose pairs annotated for spatial relationship rationality, we first construct a large-scale high-quality foreground-conditioned inpainting dataset based on pre-trained model self-generation combined with high-standard manual annotation (Sec.~\ref{subsec:datapipeline}).
Then, to avoid the gradient conflict issue in standard DPO, we propose \textbf{MaskDPO}, \yr{a spatially disentangled optimization strategy that} %which utilizes the mask of foreground subject to 
restricts the preference loss only in background region, while reusing the inpainting loss in the foreground region (Sec.~\ref{subsec:maskdpo}).
To %mitigate the incoherence at the foreground boundary introduced by MaskDPO,
\yr{enhance coherence at the foreground-background boundary,}
we propose \textbf{Conditional Asymmetric Preference Optimization}, which adjusts the position and size of foreground subject differently across win-lose pairs and performs a global preference optimization on entire images (Sec.~\ref{subsec:capo}).
In addition, to fully \yr{capture the shared commonality of plausible spatial relationships in winning samples,} %the characteristic of similar spatial information among win-samples, 
we propose \textbf{\yr{Shared Commonality Preference Optimization}}, which minimizes the \lqr{reward gap} between \yr{win-win} pairs, to %improve the model's understanding of commonalities in high-quality \yr{spatial relationships and 
\yr{further promote spatial rationality} (Sec.~\ref{subsec:psl}).

\subsection{\yr{Spatial Rationality Preference} Data %Construction
} 
\label{subsec:datapipeline}
A fundamental obstacle to \yr{applying} preference optimization techniques to foreground-conditioned inpainting is the critical scarcity of specialized training datasets, \tyz{as no public resource offers win-lose \yr{pairs} specifically annotated for the \textbf{spatial rationality} between the foreground and background.}
Since the primary issue in this task is the occurrence of spatial relationship hallucinations, constructing a high-quality task-specific dataset that provides spatial \yr{rationality} preference signals is a necessary prerequisite for training DPO-like strategies \yr{to} yield visually consistent and human-preferred results.

%address the data scarcity and task specificity, 
To \yr{this end,}
we devised a four-stage data construction pipeline \yr{for spatial rationality human preference data}. 
(1) First, we collected a large-scale \yr{set} of subject image-prompt pairs from open-source repositories, \yr{which covers} diverse common foreground subjects. 
(2) To increase the complexity and variation of the backgrounds, we employed GPT4o~\cite{achiam2023gpt} for prompt expansion, generating rich\yr{er} descriptions of the background scene. 
(3) Then we utilized the state-of-the-art foreground-conditioned inpainting model FLUX-Pinco~\cite{flux2024,lu2025pinco} to generate a pool of 20 candidate \yr{background inpainted} images for each case, ensuring a wide contrast range for preference labeling. 
(4) Finally, a rigorous human annotation and review process was implemented.
Fifty professional annotators conducted a comprehensive evaluation based on text prompts, considering spatial rationality including spatial scale, positional relationship and viewpoint matching. 
Ultimately, 2-4 winning samples and 2-4 losing samples were selected for each case.
This guarantees our dataset \yr{is} specialized for our foreground-conditioned inpainting task, providing high-contrast \yr{spatial rationality win-lose} preference signals required to optimize the model for superior spatial realism.

\subsection{MaskDPO: Spatially Disentangled Optimization}
\label{subsec:maskdpo}
\noindent\textbf{\yr{Analysis of Standard DPO on Spatial Rationality Optimization.}}
The standard DPO loss for Diffusion models~\cite{black2023training} aims to maximize the log-probability gap between a winning sample $x^w$ and a losing sample $x^l$.
This is achieved by optimizing an implicit reward $\hat{r}_\theta$ derived from the policy model $\epsilon_\theta$ and a reference model using $\epsilon_{ref}$, 
%The DPO loss is:
\yr{formulated as:}
\begin{equation}
\setlength\abovedisplayskip{4pt}
\setlength\belowdisplayskip{4pt}
\label{eq:dpo}
\small
    L_\text{DPO} = \mathbb{E}_{(c, x^w, x^l) \sim D} \left[ -\log \sigma \left( \hat{r}_\theta(x^w, c) - \hat{r}_\theta(x^l, c) \right) \right],
\end{equation}
where the implicit reward $\hat{r}_\theta$ is typically a function of the L2-norm of the predicted noise error:
\begin{equation}
\setlength\abovedisplayskip{4pt}
\setlength\belowdisplayskip{4pt}
\label{eq:r_theta}
\footnotesize
    \hat{r}_\theta(x, c) \propto \mathbb{E}_{z_t, \epsilon} \left[ \Vert \epsilon - \epsilon_{ref}(z_t, c, t) \Vert^2_2 - \Vert \epsilon - \epsilon_\theta(z_t, c, t) \Vert^2_2 \right].
\end{equation}

In foreground-conditioned inpainting, the condition $c$ refer\yr{s} to the text prompt $T$ and given foreground image, which are identical across the \yr{winning and losing} %pairs, 
\yr{samples in a human preference pair.}
%and for the sampled  win-lose pairs, their foreground regions are also completely identical.
% However, the standard DPO loss calculates the the preference optimization across the entire image.
\tyz{The standard DPO loss (Eq.~\ref{eq:dpo}) calculates the preference optimization across the entire image, which \yr{aims} to increase the probability of \yr{winning sample} $x^w$ and decrease the probability of \yr{losing sample} $x^l$.
However, since the foregrounds are identical in $x^w$ and $x^l$ \yr{in a training %human preference 
pair}, this global loss forces the policy model $\epsilon_\theta$ to learn completely \textbf{opposite gradient directions} within the foreground region. 
Specifically, the model is simultaneously pulled to increase and decrease the likelihood of generating the same \yr{foreground} latent representation.} Such conflicting gradients lead to model degradation and losing the ability to preserve foreground regions over training.

\noindent\textbf{Masked Preference Optimization.} 
To avoid the conflict, we \yr{propose Masked Preference Optimization (MPO), which performs spatially disentangled strategy} %a simple yet effective method to 
constrain\yr{ing} preference optimization only in the background region.% \yr{while applying a subject preservation loss in the foreground region}. 
We utilize the subject mask $m$ (where $m=1$ for the background) to reformulate the reward.
Specifically, we first downsample and reshape the mask $m$ to align it with the shape of $z_t$ in diffusion.
We compute the reward exclusively on the background region \tyz{by multiplying this spatially-aligned mask $m$ into the reward,} effectively bypassing optimization on the identical foreground region \yr{that causes conflicting gradients.} 
The reward $\hat{r}_\text{MPO}$ is defined as:
\begin{equation}
\setlength\abovedisplayskip{4pt}
\setlength\belowdisplayskip{4pt}
\label{eq:r_mpo}
\footnotesize
    \hat{r}_\text{MPO}(y, c) \propto \mathbb{E}_{z_t, \epsilon} \left[ \Vert (\epsilon - \epsilon_{ref}) \odot m \Vert^2_2 - \Vert (\epsilon - \epsilon_\theta) \odot m \Vert^2_2 \right].
\end{equation}
This yields our preference loss $L_\text{MPO}$ which applies DPO only to background area where distinct preference\yr{s} exist:
\begin{equation}
\setlength\abovedisplayskip{4pt}
\setlength\belowdisplayskip{4pt}
\label{eq:l_mpo}
% \small
    L_\text{MPO} = \mathbb{E} \left[ -\log \sigma \left( \hat{r}_\text{MPO}(x^w, c) - \hat{r}_\text{MPO}(x^l, c) \right) \right].
\end{equation}

\noindent\textbf{Foreground Preservation Loss.} 
While $L_{MPO}$ successfully isolates the background for preference alignment, it provides no optimization signal to maintain the \yr{consistency} of the \yr{foreground} subject, in which case the model's preservation of the foreground subject gradually decreases as training progresses.
To guarantee robust foreground preservation, we simultaneously apply the original inpainting loss, but constrain it only to the foreground region using $(1 - m)$:
\begin{equation}
\setlength\abovedisplayskip{4pt}
\setlength\belowdisplayskip{4pt}
    L_\text{Inpainting} = \mathbb{E}_{x, t, \epsilon} \left[ \Vert (\epsilon - \epsilon_\theta(z_t, c, t)) \odot (1-m) \Vert^2_2 \right].
\end{equation}

Finally, our MaskDPO combines these two components:
\begin{equation}
\label{eq:maskdpo}
\setlength\abovedisplayskip{4pt}
\setlength\belowdisplayskip{4pt}
    L_\text{MaskDPO} = L_\text{MPO} + \lambda L_\text{Inpainting},
\end{equation}
where $\lambda$ is a hyperparameter balancing background preference alignment and foreground preservation. 
This \yr{spatially disentangled optimization} mechanism %forces the model to learn human-preferences-based spatial rationality in the background via $L_\text{MPO}$, while explicitly enforcing high-fidelity subject preservation in the foreground via $L_\text{Inpainting}$.
%This spatially disentangled optimization mechanism 
\yr{enables MaskDPO to refine background spatial relationships without compromising the preservation of the foreground subject.}

\begin{figure}
    \centering
    \vspace{-0.1in}
    \includegraphics[width=0.94\linewidth]{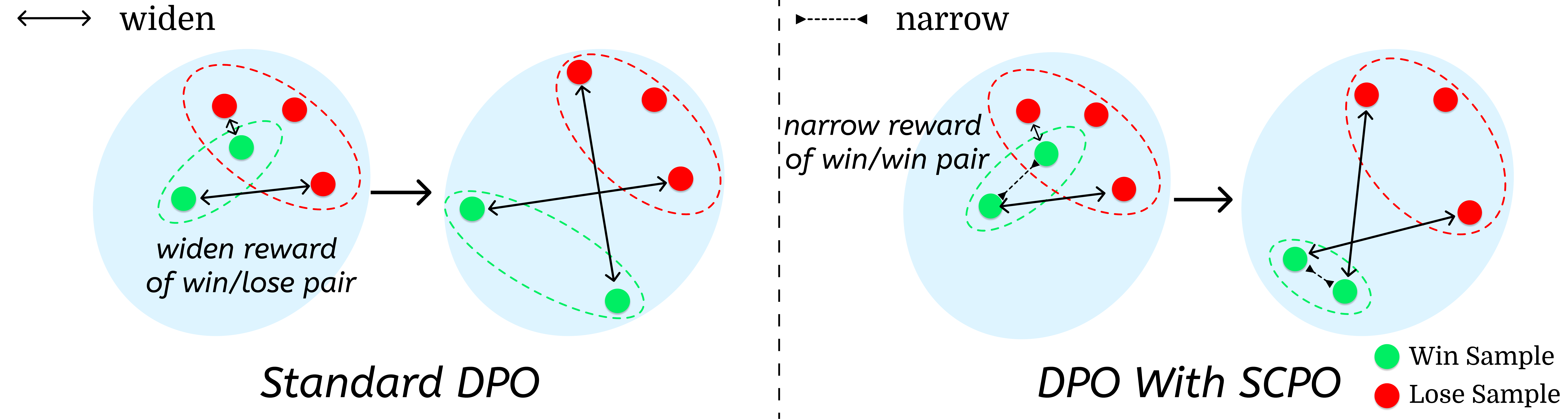}
    \vspace{-0.15cm}
    \caption{Comparison of Standard DPO and DPO with SCPO.
    }
    \label{fig:scpo}
    \vspace{-0.5cm}
\end{figure}

\subsection{Conditional Asymmetric Preference Optimization}
\label{subsec:capo}

\yr{While MaskDPO effectively improves rational spatial relationships within the background, its disentangled optimization does not explicitly address how the background interacts with foreground at boundary.} Hence, we propose \textbf{Conditional Asymmetric Preference Optimization (CAPO)} \yr{to specialize in this interaction, ensuring a coherent and natural transition at the foreground-background boundary.}

%We revert to the full-image preference optimization formulation, but eliminate the gradient conflict by introducing 
\yr{CAPO re-introduces a global preference optimization objective, but prevents gradient conflicts through}
a \tyz{Differentiated Cropping Operation}.
\yr{By ensuring}
%if 
the \yr{positions of foreground} subjects \yr{in winning and losing samples} are not perfectly aligned pixel-\yr{by}-pixel, %they will not cause 
\yr{our approach inherently prevents}
the gradient conflict observed in standard DPO\yr{, enabling stable full-image preference optimization with superior boundary coherence.}
Specifically, given a win\yr{-}lose pair $(\boldsymbol{x}_{0}^{w}$, $\boldsymbol{x}_{0}^{l})$, we apply different cropping operation\yr{s} to obtain $\boldsymbol{\widehat{{x}_0^{w}}}$ and $\boldsymbol{\widehat{{x}_0^{l}}}$.
This operation is designed to preserve subject integrity while deliberately assigning the subject to distinct relative spatial positions in the two images. Concurrently, we impose a Conditional Asymmetric Preference Optimization loss on the resulting $\boldsymbol{\widehat{{x}_0^{w}}}$ and $\boldsymbol{\widehat{{x}_0^{l}}}$ pairs:
\begin{equation}
\setlength\abovedisplayskip{4pt}
\setlength\belowdisplayskip{4pt}
\footnotesize
    L_{\text{CAPO}} = \mathbb{E}_{(c, \widehat{x^w}, \widehat{x^l}) \sim D} \left[ -\log \sigma \left( \hat{r}_\theta(\widehat{x^w}, T) - \hat{r}_\theta(\widehat{x^l}, T) \right) \right].
\end{equation}
\tyz{Such \yr{a} Differentiated Cropping Operation can eliminate the gradient conflicts \yr{in standard DPO}, while promoting the contextual awareness at the foreground boundary, thereby \yr{effectively enhancing boundary coherence.}
}

\begin{figure*}[t]
    \centering
    \vspace{-0.2in}
    \includegraphics[width=0.95\textwidth]{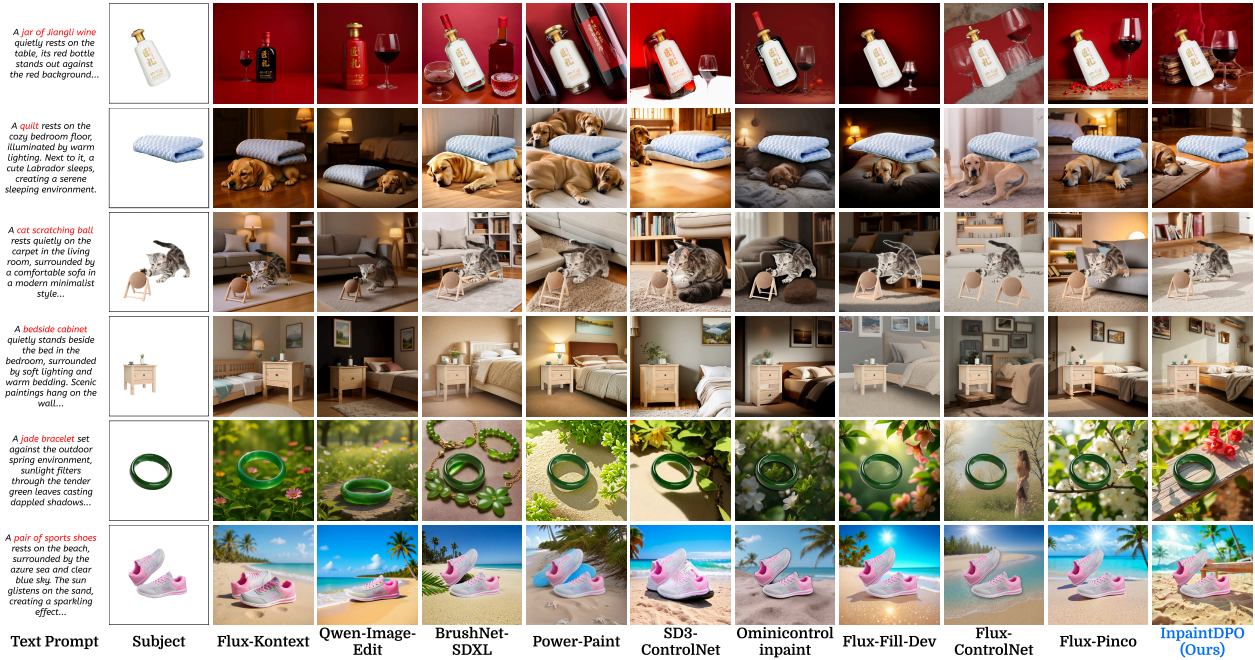}
    \vspace{-0.1in}
    \caption{Qualitative comparison of existing inpainting or editing methods and our \textit{InpaintDPO}. Our method generates plausible spatially rational images without hallucinations like inappropriate scale or position, with great prompt alignment and foreground consistency.}
    \label{fig: vision comp}
    \vspace{-0.17in}
\end{figure*}

\subsection{Shared Commonality Preference Optimization}
\label{subsec:psl}
\tyz{In our task, we observe that winning samples generated under \yr{the same} conditions often share a significant \textbf{commonality in their spatial layout and positional relationship}.
This suggests that favorable foreground-conditioned denoising trajectories in the latent space tend to exhibit a convergence around reasonable placements. \yr{Based on this observation, we propose \textbf{Shared Commonality Preference Optimization (SCPO)} to explicitly guide the model to capture this positional relationship commonality.
\tyz{SCPO}} encourages the model \yr{to} narrow the implicit reward gap between win-win pairs\yr{, assigning high and close rewards for different winning samples, thereby reinforcing the shared and plausible spatial characteristics.}}

\tyz{
Different from DPO which aims to maximize the likelihood % that the model assigns a higher implicit reward to the winning sample $x^w$ than to the losing sample $x^l$,
gap between winning samples $x^w$ and losing samples $x^l$,
the core idea of \tyz{SCPO} is to minimize the divergence of implicit rewards between two winning samples to assign similar values to both, as \yr{shown} in Fig.~\ref{fig:scpo}. 
Specifically, \tyz{SCPO} is built upon the introduction of the win-win pair, consisting of two distinct human-preferred generated images ($\boldsymbol{x}^{w_1}, \boldsymbol{x}^{w_2}$) under the same condition $\boldsymbol{c}$. % of foreground image and text prompt.
We define the \tyz{SCPO} loss by measuring the absolute difference between the implicit rewards of a paired high-quality winning samples $(c, x^w_1, x^w_2) \in \mathcal{D}'$, where $\mathcal{D}'$ is a dataset of win-win pairs. 
The loss is formulated using a symmetric, non-linear penalty function (detailed derivation in Suppl):
\begin{equation}
\setlength\abovedisplayskip{4pt}
\setlength\belowdisplayskip{4pt}
\label{eq:psl}
\scriptstyle
    L_{\text{SCPO}} = \mathbb{E}_{(c, x^{w_1}, x^{w_2}) \sim D'} \left[ -\log \left( 1 - \sigma \left( \left| \hat{r}_\theta(x^{w_1}, c) - \hat{r}_\theta(x^{w_2}, c) \right| \right) \right) \right],
\end{equation}
where the implicit reward $\hat{r}_\theta$ is \yr{the} same as Eq.~\ref{eq:r_theta}.
By narrowing the gap of implicit rewards, \yr{the $L_{\text{SCPO}}$} effectively promotes the model's understanding of positional commonality within the win-win pair, leading to improved perceptual quality and structural consistency in generated results.
Also, it prevents the over-optimization of a singular winning sample and encourages a stable convergence of implicit reward over the winning data distribution.
}

\tyz{Finally, our training loss is a weighted sum of \yr{the} above three losses:
\begin{equation}
\label{final loss}
\setlength\abovedisplayskip{4pt}
\setlength\belowdisplayskip{4pt}
    L = L_\text{MaskDPO} + \gamma L_\text{CAPO} + \mu L_\text{SCPO},
\end{equation}
where $\gamma$ and $\mu$ are hyperparameters balancing the Loss terms $L_\text{CAPO}$ and $L_\text{SCPO}$ 
respectively.
}

\section{Experiments}

\begin{table*}[t]
\centering
\vspace{-0.1in}
\caption{Quantitative Comparisons \yr{with state-of-the-art inpainting methods}. \yr{Best results are in \textcolor{red}{red}, second-best results are in \textcolor{blue}{blue}.}}
\label{tab: method compare}
\renewcommand{\arraystretch}{0.7}
\vspace{-0.05in}
\setlength\tabcolsep{2pt}
\scalebox{0.8}{
\begin{tabular}{c|c|cc|ccc|cc|cc}
\toprule
\multirow{2}*{Task} & \multirow{2}{*}{\diagbox{Methods}{Metrics}} & \multicolumn{2}{c|}{Foreground Consistency} & \multicolumn{3}{c|}{Spatial Rationality} & \multicolumn{2}{c|}{Text-alignment} & \multicolumn{2}{c}{Image Quality} \\
& & OER(BiRefNet) $\downarrow$ & OER(SAM2.1)$\downarrow$ & Scale$\uparrow$ & Position$\uparrow$ & Viewpoint$\uparrow$ & VQAScore $\uparrow$ & FV2Score $\downarrow$ & {FID $\downarrow$} & Context$\uparrow$ \\
\midrule
\multicolumn{1}{c|}{\multirow{2}*{Edit}} & Flux-Kontext~\cite{labs2025fluxkontext} & 40.7\% & 44.0\% & 3.84 & 3.88 & 3.92 & \best{0.884} & 22.3\% & 111.43 & 0.159 \\
& Qwen-Image-Edit~\cite{wu2025qwen} & 34.8\% & 23.7\% & \second{3.89} & \second{3.95} & \second{3.96} & \second{0.876} & 22.2\% & 115.72 & \second{0.254} \\
\midrule
UNet-based & PowerPaint~\cite{zhuang2023powerpaint} & 27.4\% & 21.9\% & 3.44 & 3.61 & 3.84 & 0.744 & 12.7\% & 111.78 & 0.242 \\
Inpainting & BrushNet~\cite{ju2024brushnet} & 18.3\% & 15.4\% & 3.57 & 3.71 & 3.46 & 0.826 & 9.4\% & 107.76 & 0.262 \\
\midrule
\multicolumn{1}{c|}{\multirow{6}*{\makecell{MMDiT-based \\ Inpainting}}} & SD3-ControlNet~\cite{alimama_SD3} & 22.3\% & 8.7\% & 3.72 & 3.68 & 3.78 & 0.873 & 12.6\% & 103.91 & 0.248 \\
& Ominicontrol-inpaint~\cite{tan2025ominicontrol} & 24.2\% & 8.4\% & 3.53 & 3.68 & 3.71 & 0.833 & 3.5\% &  117.29 & \second{0.254} \\
& Flux-Fill-dev~\cite{flux2024inpaint} & 12.3\% & 6.1\% & 3.74 & 3.48 & 3.56 & 0.820 & 9.0\% & 115.30 & 0.238 \\
& Flux-ControlNet~\cite{alimama_FLUX} & 16.7\% & 9.0\% & 3.63 & 3.72 & 3.80 & 0.870 & \second{3.3\%} & 119.44 & 0.245 \\
& Flux-Pinco~\cite{lu2025pinco} & \second{8.4\%} & \second{4.3\%} & 3.87 & 3.91 & 3.92 & 0.842 & 5.3\% & \second{99.05} & 0.249 \\
& \textbf{InpaintDPO} & \best{5.1\%} & \best{2.7\%} & \best{3.94} & \best{4.05} & \best{4.02} & \second{0.876} & \best{1.2\%} & \best{96.43} & \best{0.268} \\
\bottomrule
\end{tabular}
}
\vspace{-0.4cm}
\end{table*}

\subsection{Experiment Settings}
\lqr{\noindent\textbf{Dataset.} The dataset for training our InpaintDPO was meticulously constructed, as detailed in Sec.~\ref{subsec:datapipeline}. We began by collecting 3k high-quality image-prompt pairs from online inpainting data, reserving 500 pairs for evaluation. Subsequently, we generated masks using BiRefNet~\cite{zheng2024bilateral} and enriched the text prompts with GPT-4o~\cite{achiam2023gpt}. Next, we generated 1k-resolution images across various aspect ratios (1:1, 9:16, and 16:9) using FLUX-Pinco~\cite{lu2025pinco,flux2024}. Finally, 50 professional annotators performed rigorous preference labeling, primarily focusing on spatial rationality. This process resulted in a comprehensive preference dataset containing over 10k win-lose pairs and 5k win-win pairs.}

\noindent\textbf{Implementation Details.}
We conduct our training on finetuning the FLUX-Pinco~\cite{lu2025pinco,flux2024} using LoRA~\cite{hu2022lora,tian2024hydralora} (detailed architecture in Suppl).
Our training is conducted on 8 NVIDIA 96G H20 GPUs for 5 epochs with a batch size of 2. During training, we adopt AdamW~\cite{loshchilov2017decoupled} optimizer and the learning rate was maintained at 0.0001, using a 1K-iteration warmup. In Eq. \ref{eq:maskdpo} and \ref{final loss}, the hyperparameters $\lambda,\gamma,\mu$ are set to 2, 1 and 0.5 respectively.

\noindent\textbf{Evaluation Metrics.}
Following Pinco~\cite{lu2025pinco}, we comprehensively evaluate the performance of our model %using five key metrics: 
\yr{from four aspects:}
Image Quality, Foreground Consistency, Text-Alignment, and Spatial Rationality.
(1) Image Quality: We utilize the Fréchet Inception Distance (FID)~\cite{heusel2017gans} on MSCOCO~\cite{lin2014microsoft} to assess the overall realism and quality of the generated images.
Also we follow OmniPaint~\cite{yu2025omnipaint} to evaluate the Context Coherence Score based on DINOv2~\cite{oquab2023dinov2}.
(2) Foreground consistency: To quantitatively measure the preservation and consistency of the foreground subject, we employ BiRefNet~\cite{zheng2024bilateral} and SAM2.1~\cite{ravi2024sam2} for accurate subject mask generation and then calculate the Object Extension Ratio (OER)~\cite{chen2024virtualmodel} to evaluate the
consistency between the subject shape in the generated image and the ground truth subject shape.
(3) Text alignment: we use VQA Score~\cite{lin2025evaluating} \yr{and FV2Score~\cite{xiao2024florence}} to assess the subject-text alignment and the multi-subject rate.
(4) \yr{Spatial} rationality: we employ the multi-modal large language model GPT-4o~\cite{achiam2023gpt} for an automated assessment of the subject's positional rationality based on three aspects: spatial size, positional relationship, and viewpoint. 
Each aspect is scored on a scale of 1, 3, or 5. 

\begin{table}[t]
  \centering
  \setlength\tabcolsep{4pt}
  \tiny
  % \vspace{-0.1in}
  \caption{\yr{User Study Results:} ELO Ranking.}
  \vspace{-0.1in}
  \begin{tabular}{c l c c | c l c c}
    \toprule
    Rank & Method & ELO$\uparrow$ & \# Appearances & Rank & Method & ELO$\uparrow$ & \# Appearances \\
    \midrule
    1 & \textbf{InapintDPO} & 1432 & 2850 & 5 & PowerPaint & 871 & 2570 \\
    2 & Pinco & 1332 & 2570 & 6 & Flux-Fill-dev & 853 & 2180 \\
    3 & BrushNet & 1115 & 2310 & 7 & OminiControl & 784 & 2490 \\
    4 & Flux-ControlNet & 926 & 2110 & 8 & SD3-ControlNet & 686 & 2500 \\
    \bottomrule
  \end{tabular}
  \vspace{-0.2in}
  \label{tab:elo_rank}
\end{table}

\subsection{Comparisons with Existing Methods}
We conduct qualitative and quantitative comparisons, along with a user study, to demonstrate the superiority of our InpaintDPO.
We compare with UNet-based inpainting methods, including PowerPaint~\cite{zhuang2023powerpaint} and BrushNet~\cite{ju2024brushnet}, %CTR-Driven~\cite{chen2025ctr} 
and MMDiT-based inpainting methods including SD3-ControlNet~\cite{alimama_SD3}, Flux-Fill-dev~\cite{flux2024inpaint}, Flux-ControlNet~\cite{alimama_FLUX}, Flux-OminiControl~\cite{tan2025ominicontrol} and Flux-Pinco~\cite{lu2025pinco}
Additionally, we compare with two unified image generation and editing models, FLUX.1 Kontext~\cite{labs2025fluxkontext} and Qwen Image~\cite{wu2025qwen}, where we ask the model to perform background replacement while maintaining complete consistency of the input subject.

\noindent\textbf{Qualitative Results.} 
Fig.~\ref{fig: vision comp} shows the qualitative comparisons with previous inpainting and editing methods.
In foreground-conditioned inpainting tasks, certain models face challenges in seamlessly integrating the subject with the background specified by prompts, as illustrated in the 2nd, 3rd, and 5th rows, leading to incorrect layout arrangements and object interaction.
Furthermore, some methods struggle with subject expansion, a difficulty notably observed in the bedside table of the 4th row. 
For special subjects requiring multiple surfaces for support, as in 1st and 6th rows, several models find it problematic to accurately process this information, resulting in incorrect generation of just one single support surface.
Notably, for editing methods, they possess fewer spatial relationship hallucinations, but suffer from \yr{a} severe problem of foreground subject preservation, visibly deforming or changing the subjects required in inpainting task.

\begin{table}[t]
\centering
\caption{Quantitative \yr{Ablation Studies on} %for the Text-Guided Subject-Position Variable Background Inpainting task.
\yr{Core Components: MaskDPO, CAPO, and SCPO.}
}
\label{tab: module ablation}
\vspace{-0.05in}
\setlength\tabcolsep{2pt}
\scalebox{0.45}{
\begin{tabular}{ccc|cc|ccc|c|cc} 
\toprule
\multicolumn{3}{c|}{\textbf{Methods}} & \multicolumn{2}{c|}{\textbf{Foreground Consistency}} & \multicolumn{3}{c|}{\textbf{Spatial Rationality}} & \textbf{Text-alignment} & \multicolumn{2}{c}{\textbf{Image Quality}} \\
MaskDPO & CAPO & \yr{SCPO} & \textbf{OER(BiRefNet)} $\downarrow$ & \textbf{OER(SAM2.1)}$\downarrow$ & \textbf{Scale}$\uparrow$ & \textbf{Position}$\uparrow$ & \textbf{Viewpoint}$\uparrow$ & \textbf{VQA Score} $\uparrow$ & {\textbf{FID} $\downarrow$} & \textbf{Context}$\uparrow$ \\
\midrule
& & & 8.4\% & 4.3\% & 3.87 & 3.91 & 3.92 & 0.842 & 99.05 & 0.249 \\
\checkmark & & & 5.6\% & 3.4\% & 3.92 & 3.94 & 3.94 & 0.863 & 98.87 & 0.250 \\
\checkmark & \checkmark & & 5.6\% & 3.1\% & 3.90 & 3.96 & 3.92 & 0.866 & 97.42 & 0.258 \\
\checkmark & \checkmark & \checkmark & \textbf{5.1\%} & \textbf{2.7\%} & \textbf{3.94} & \textbf{4.05} & \textbf{4.02} & \textbf{0.876} & \textbf{96.43} & \textbf{0.268} \\
\bottomrule
\end{tabular}
}
\vspace{-0.3cm}
\end{table}

\begin{figure}[t]
    \centering
    \includegraphics[width=\linewidth]{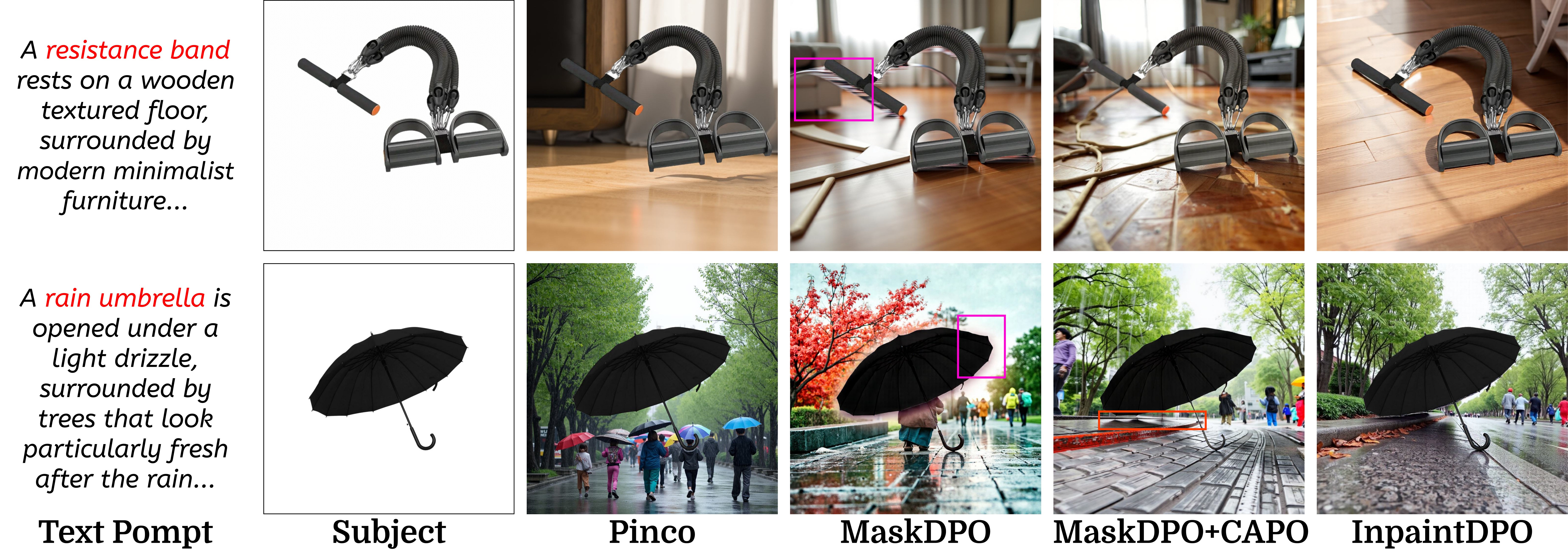}
    \vspace{-0.1in}
    \caption{Qualitative Ablation %Comparisons 
    \yr{Studies on} InpaintDPO's Core Components.}
    \label{fig: ablation core component}
    \vspace{-0.15in}
\end{figure}

\noindent\textbf{Quantitative Results.}
As presented in Table \ref{tab: method compare}, our InpaintDPO achieves leading performance across various evaluation metrics.
Regarding Image Quality, we achieve the best results, demonstrated by the lowest FID for high realism and the highest Context Coherence Score for superior contextual integration.
For Foreground Consistency, our model outperforms in both OER scores, validating precise foreground preservation.
Also, our model achieves the best spatial rationality in all three aspects, proving its enhanced ability to generate visually and spatially plausible inpainting results.
In terms of Text Alignment, InpaintDPO achieves lowest FV2Score and the highest VQAScore among inpainting models, being second only to editing models, which do trade off between subject preservation and image quality (as shown in Tab.~\ref{tab: method compare} and Fig.~\ref{fig: vision comp}, which is unacceptable for inpainting tasks).

\noindent\textbf{User Study.}
\lqr{We conducted a large-scale AI Inpainting Arena which employs the ELO rating system~\cite{elo1978rating} to measure human preference to evaluate spatial rationality and perceptual quality (detailed setting in Suppl).}
We involve all the inpainting methods and invite 30 experts to choose images with better spatial rationality. 
After collecting over 10k votes, InpaintDPO achieved the highest ELO score, establishing a considerable lead over other models as in Tab.~\ref{tab:elo_rank}.

\subsection{Ablation Study}

\noindent\textbf{\yr{Core Components.}}
To investigate the effectiveness and individual contributions of the proposed modules MaskDPO, CAPO and SCPO, we conduct a comprehensive ablation study based on progressive additions, as detailed in Tab.~\ref{tab: module ablation} and Fig.~\ref{fig: ablation core component}.
The introduction of MaskDPO yields the most significant improvement, particularly in \yr{improving spatial rationality (Fig.~\ref{fig: ablation core component}), and} ensuring foreground consistency, reducing the $\text{OER}(\text{BiRefNet})$ substantially from $8.4\%$ to $5.6\%$, validating the necessity of background preference alignment.
Following this, integrating the CAPO further enhances the foreground boundary coherence, especially evidenced by improvements in Fig.~\ref{fig: ablation core component} and perceptual metrics of Context score.
The final inclusion of the SCPO introduces a further \yr{spatial rationality} improvement, achieving the best overall performance across all quality metrics, proving the necessity and complementary nature of each module for advancing in foreground-conditioned inpainting.

\begin{table}[t]
\centering
\vspace{-0.2in}
\caption{Quantitative \yr{Ablation Studies on} %Comparisons for the Text-Guided Subject-Position Variable Background Inpainting task.
\yr{subject preservation in MaskDPO.}
}
\label{tab: maskdpo}
\vspace{-0.1in}
\setlength\tabcolsep{2pt}
\scalebox{0.45}{
\begin{tabular}{ll|cc|ccc|c|cc}
\toprule
\multicolumn{2}{c|}{\multirow{2}{*}{\textbf{Combination}}} & \multicolumn{2}{c|}{\textbf{Foreground Consistency}} & \multicolumn{3}{c|}{\textbf{Spatial Rationality}} & \textbf{Text-alignment} & \multicolumn{2}{c}{\textbf{Image Quality}} \\
\multicolumn{2}{c|}{} & \textbf{OER(BiRefNet)} $\downarrow$ & \textbf{OER(SAM2.1)}$\downarrow$ & \textbf{Scale}$\uparrow$ & \textbf{Position}$\uparrow$ & \textbf{Viewpoint}$\uparrow$ & \textbf{VQA Score} $\uparrow$ & {\textbf{FID} $\downarrow$} & \textbf{Context}$\uparrow$ \\
\midrule
\multicolumn{2}{c|}{\text{Standard DPO}} & 8.4\% & 4.3\% & 3.52 & 3.44 & 3.72 & 0.817 & 115.37 & 0.232 \\
\multicolumn{2}{c|}{$L_{\text{MPO}}+L_\text{Subject-\yr{SCPO}}$} & 7.8\% & 4.1\% & \textbf{3.93} & 3.92 & \textbf{3.94} & 0.858 & 99.37 & 0.246 \\
\multicolumn{2}{c|}{$L_{\text{MPO}}+L_\text{Inpainting}$} & \textbf{5.6\%} & \textbf{3.4\%} & 3.92 & \textbf{3.94} & \textbf{3.94} & \textbf{0.863} & \textbf{98.87} & \textbf{0.250} \\
\bottomrule
\end{tabular}
}
\vspace{-0.1in}
\end{table}

\begin{figure}[t]
    \centering
    \vspace{-0.05in}
    \includegraphics[width=\linewidth]{figs/Ablation_foreground-18.pdf}
    \vspace{-0.2in}
    \caption{Qualitative Ablation Comparisons of Pinco, \yr{standard} DPO, MPO+Subject SCPO Loss and MaskDPO (Ours).}
    \label{fig: ablation foreground}
    \vspace{-0.2in}
\end{figure}

\noindent\textbf{Subject Preservation in MaskDPO.}
Our MaskDPO addresses the conflicting gradients in foreground region caused in Standard DPO \yr{by spatially disentangled optimization: background preference optimization and foreground inpainting loss}.
\yr{To validate its design, we compare MaskDPO with two alternative strategies: 1) Standard DPO, and 2) replacing the foreground inpainting loss $L_\text{Inpainting}$ with SCPO, minimizing the implicit reward within forground for subject preservation. As shown in}
Fig.~\ref{fig: ablation foreground} \yr{and Tab.~\ref{tab: maskdpo}}, 
Standard DPO can hardly preserve the subject.
\yr{Replacing $L_\text{Inpainting}$ with foreground SCPO} \yr{sometimes suffers from subject extension} and color differences. In contrast, our MaskDPO \yr{effectively} addresses these limitations, \yr{well} preserves subject details and suppresses subject expansion, making it a more effective solution for subject preservation, which is \yr{crucial} for high-quality inpainting tasks.

\subsection{Method Migration}
\begin{figure}[t]
    \centering
    \vspace{-0.1in}
    \includegraphics[width=0.9\linewidth]{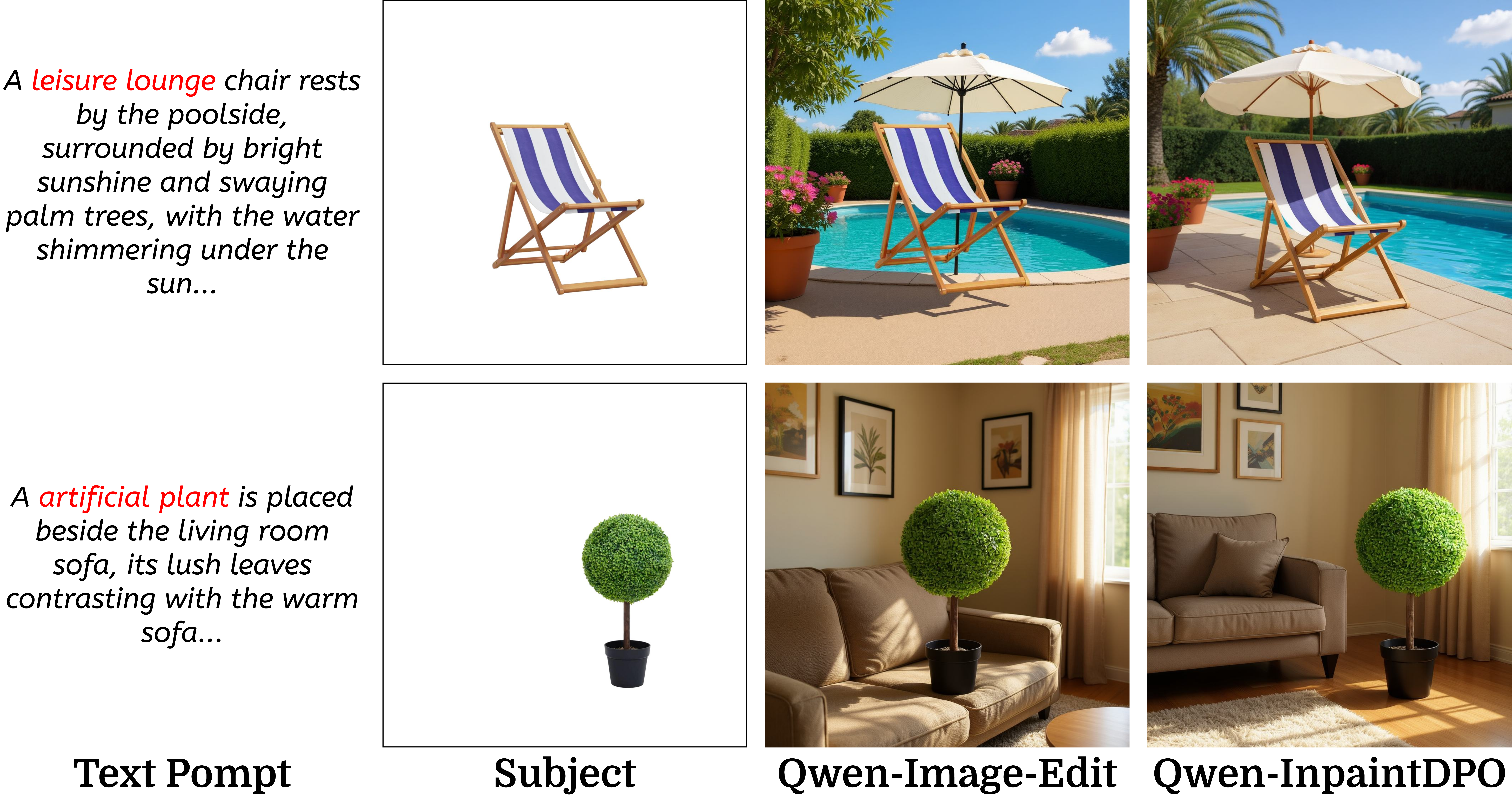}
    \vspace{-0.1in}
    \caption{Qualitative Comparisons \yr{between} Qwen-Image-Edit and Qwen-InapintDPO.}
    \label{fig: qwendpo}
    \vspace{-0.2in}
\end{figure}

To validate the robustness and generalizability of our proposed InpaintDPO, we further conduct a transferability experiment by applying our InpaintDPO to an entirely different base model: the Qwen-Image-edit~\cite{wu2025qwen} (detailed settings in Suppl).
Specifically, as \yr{shown} in Tab.~\ref{tab: qwen dpo} and Fig.~\ref{fig: qwendpo}, applying InpaintDPO successfully enabled the image editing model to achieve \lqr{better subject detail preservation and improved spatial rationality.}
It demonstrates that our InpaintDPO is a generic and effective strategy for mitigating the Spatial Relationship Hallucinations while preserving foreground subjects in foreground-conditioned inpainting task.

\begin{table}[t]
\vspace{-0.2in}
\centering
\caption{Quantitative Comparisons for \yr{method migration on Qwen-Image-editing.}
}
\label{tab: qwen dpo}
\vspace{-0.1in}
\setlength\tabcolsep{2pt}
\scalebox{0.45}{
\begin{tabular}{cc|cc|ccc|c|cc}
\toprule
\multicolumn{2}{c|}{\multirow{2}{*}{\textbf{Method}}} & \multicolumn{2}{c|}{\textbf{Foreground Consistency}} & \multicolumn{3}{c|}{\textbf{Spatial Rationality}} & \textbf{Text-alignment} & \multicolumn{2}{c}{\textbf{Image Quality}} \\
\multicolumn{2}{c|}{} & \textbf{OER(BiRefNet)} $\downarrow$ & \textbf{OER(SAM2.1)}$\downarrow$ & \textbf{Scale}$\uparrow$ & \textbf{Position}$\uparrow$ & \textbf{Viewpoint}$\uparrow$ & \textbf{VQA Score} $\uparrow$ & {\textbf{FID} $\downarrow$} & \textbf{Context}$\uparrow$ \\
\midrule
\multicolumn{2}{c|}{\text{Qwen-Image-Edit}} & 34.8\% & 23.7\% & 3.89 & 3.95 & 3.96 & 0.876 & 115.72 & 0.254 \\
\multicolumn{2}{c|}{\text{Qwen-Inpaint DPO}} & \textbf{13.7\%} & \textbf{9.4\%} & \textbf{4.03} & \textbf{3.97} & \textbf{3.96} & \textbf{0.878} & \textbf{105.19} & \textbf{0.262} \\
\bottomrule
\end{tabular}
}

\end{table}

\section{Conclusion}
\label{sec:conclusion}
In this paper, we address the prevalent and challenging Spatial Relationship Hallucinations in foreground-conditioned inpainting, where existing models suffer from inappropriate spatial scale, positional relationship, and viewpoint between the given subject and generated background. 
Due to the subjectivity and poor quantifiability of spatial rationality, we propose InpaintDPO, a novel post-training framework dedicated to enhancing spatial rationality for this task via diverse preference optimization strategies. 
To resolve critical gradient conflicts of standard DPO in this scenario, we first introduce MaskDPO, which uses the foreground mask to confine preference optimization exclusively to the background while robustly preserving the foreground via inpainting loss. 
We further enhance foreground boundary coherence through Conditional Asymmetric Preference Optimization, which employs differentiated cropping operation to boost contextual awareness. 
Finally, we propose Shared Commonality Preference Optimization to improve the model’s understanding of spatial commonalities across winning samples. 
Extensive experiments demonstrate that InpaintDPO significantly mitigates Spatial Relationship Hallucinations, achieving superior spatial rationality and visual quality compared to existing methods. 
Additionally, the successful transferability on Qwen-Image-edit validates its broad adaptability across different architectures.
{
    \small
    \bibliographystyle{ieeenat_fullname}
    \bibliography{main}

@String(TOG= {ACM Trans. Graph.})

@String(ICLR = {Int. Conf. Learn. Represent.})

@String(AAAI = {AAAI})

@String(TOG   = {ACM TOG})

@String(ICLR  = {ICLR})

@inproceedings{ldm,
  title={High-resolution image synthesis with latent diffusion models},
  author={Rombach, Robin and Blattmann, Andreas and Lorenz, Dominik and Esser, Patrick and Ommer, Bj{\"o}rn},
  booktitle={Proceedings of the IEEE/CVF conference on computer vision and pattern recognition},
  pages={10684--10695},
  year={2022}
}

@article{sdxl,
  title={Sdxl: Improving latent diffusion models for high-resolution image synthesis},
  author={Podell, Dustin and English, Zion and Lacey, Kyle and Blattmann, Andreas and Dockhorn, Tim and M{\"u}ller, Jonas and Penna, Joe and Rombach, Robin},
  journal={arXiv preprint arXiv:2307.01952},
  year={2023}
}

@inproceedings{dit,
  title={Scalable diffusion models with transformers},
  author={Peebles, William and Xie, Saining},
  booktitle={Proceedings of the IEEE/CVF International Conference on Computer Vision},
  pages={4195--4205},
  year={2023}
}

@inproceedings{sd3,
  title={Scaling rectified flow transformers for high-resolution image synthesis},
  author={Esser, Patrick and Kulal, Sumith and Blattmann, Andreas and Entezari, Rahim and M{\"u}ller, Jonas and Saini, Harry and Levi, Yam and Lorenz, Dominik and Sauer, Axel and Boesel, Frederic and others},
  booktitle={Forty-first International Conference on Machine Learning},
  year={2024}
}

@misc{flux2024,
    author={Black Forest Labs},
    title={FLUX},
    year={2024},
    howpublished={\url{https://github.com/black-forest-labs/flux}},
}

@article{chen2023pixart,
  title={Pixart-alpha: Fast training of diffusion transformer for photorealistic text-to-image synthesis},
  author={Chen, Junsong and Yu, Jincheng and Ge, Chongjian and Yao, Lewei and Xie, Enze and Wu, Yue and Wang, Zhongdao and Kwok, James and Luo, Ping and Lu, Huchuan and others},
  journal={arXiv preprint arXiv:2310.00426},
  year={2023}
}

@article{ju2024brushnet,
  title={Brushnet: A plug-and-play image inpainting model with decomposed dual-branch diffusion},
  author={Ju, Xuan and Liu, Xian and Wang, Xintao and Bian, Yuxuan and Shan, Ying and Xu, Qiang},
  journal={arXiv preprint arXiv:2403.06976},
  year={2024}
}

@article{zhuang2023powerpaint,
  title={A task is worth one word: Learning with task prompts for high-quality versatile image inpainting},
  author={Zhuang, Junhao and Zeng, Yanhong and Liu, Wenran and Yuan, Chun and Chen, Kai},
  journal={arXiv preprint arXiv:2312.03594},
  year={2023}
}

@inproceedings{controlnet,
  title={Adding conditional control to text-to-image diffusion models},
  author={Zhang, Lvmin and Rao, Anyi and Agrawala, Maneesh},
  booktitle={Proceedings of the IEEE/CVF International Conference on Computer Vision},
  pages={3836--3847},
  year={2023}
}

@article{manukyan2023hd,
  title={HD-Painter: high-resolution and prompt-faithful text-guided image inpainting with diffusion models},
  author={Manukyan, Hayk and Sargsyan, Andranik and Atanyan, Barsegh and Wang, Zhangyang and Navasardyan, Shant and Shi, Humphrey},
  journal={arXiv preprint arXiv:2312.14091},
  year={2023}
}

@article{avrahami2023blended,
  title={Blended latent diffusion},
  author={Avrahami, Omri and Fried, Ohad and Lischinski, Dani},
  journal={ACM transactions on graphics (TOG)},
  volume={42},
  number={4},
  pages={1--11},
  year={2023},
  publisher={ACM New York, NY, USA}
}

@inproceedings{eshratifar2024salient,
  title={Salient Object-Aware Background Generation using Text-Guided Diffusion Models},
  author={Eshratifar, Amir Erfan and Soares, Joao VB and Thadani, Kapil and Mishra, Shaunak and Kuznetsov, Mikhail and Ku, Yueh-Ning and De Juan, Paloma},
  booktitle={Proceedings of the IEEE/CVF Conference on Computer Vision and Pattern Recognition},
  pages={7489--7499},
  year={2024}
}

@inproceedings{lu2025pinco,
  title={Pinco: Position-induced consistent adapter for diffusion transformer in foreground-conditioned inpainting},
  author={Lu, Guangben and Du, Yuzhen and Tang, Yizhe and Sun, Zhimin and Yi, Ran and Qi, Yifan and Wang, Tianyi and Ma, Lizhuang and Zou, Fangyuan},
  booktitle={Proceedings of the IEEE/CVF International Conference on Computer Vision},
  pages={15266--15276},
  year={2025}
}

@inproceedings{avrahami2022blended,
  title={Blended diffusion for text-driven editing of natural images},
  author={Avrahami, Omri and Lischinski, Dani and Fried, Ohad},
  booktitle={Proceedings of the IEEE/CVF conference on computer vision and pattern recognition},
  pages={18208--18218},
  year={2022}
}

@inproceedings{corneanu2024latentpaint,
  title={Latentpaint: Image inpainting in latent space with diffusion models},
  author={Corneanu, Ciprian and Gadde, Raghudeep and Martinez, Aleix M},
  booktitle={Proceedings of the IEEE/CVF Winter Conference on Applications of Computer Vision},
  pages={4334--4343},
  year={2024}
}

@inproceedings{wang2025towards,
  title={Towards Enhanced Image Inpainting: Mitigating Unwanted Object Insertion and Preserving Color Consistency},
  author={Wang, Yikai and Cao, Chenjie and Yu, Junqiu and Fan, Ke and Xue, Xiangyang and Fu, Yanwei},
  booktitle={Proceedings of the Computer Vision and Pattern Recognition Conference},
  pages={23237--23248},
  year={2025}
}

@inproceedings{tan2025ominicontrol,
  title={Ominicontrol: Minimal and universal control for diffusion transformer},
  author={Tan, Zhenxiong and Liu, Songhua and Yang, Xingyi and Xue, Qiaochu and Wang, Xinchao},
  booktitle={Proceedings of the IEEE/CVF International Conference on Computer Vision},
  pages={14940--14950},
  year={2025}
}

@inproceedings{hu2025high,
  title={High-efficient diffusion model fine-tuning with progressive sparse low-rank adaptation},
  author={Hu, Teng and Zhang, Jiangning and Yi, Ran and Huang, Hongrui and Wang, Yabiao and Ma, Lizhuang},
  booktitle={13th International Conference on Learning Representations, ICLR 2025},
  pages={92066--92078},
  year={2025},
  organization={International Conference on Learning Representations, ICLR}
}

@misc{flux2024inpaint,
    author={Black Forest Labs},
    title={FLUX-Fill-dev},
    year={2024},
    howpublished={\url{https://github.com/black-forest-labs/flux/blob/main/docs/fill.md}},
}

@misc{alimama_FLUX,
  author={AliMama Creative},
  title={FLUX.1-dev Controlnet Inpainting Beta},
  year={2024},
  howpublished= {\url{https://huggingface.co/alimama-creative/FLUX.1-dev-Controlnet-Inpainting-Beta}},
}

@misc{alimama_SD3,
  author={AliMama Creative},
  title={SD3 Controlnet Inpainting},
  year={2024},
  howpublished= {\url{https://huggingface.co/alimama-creative/SD3-Controlnet-Inpainting}},
}

@article{he2024freeedit,
  title={FreeEdit: Mask-free Reference-based Image Editing with Multi-modal Instruction},
  author={He, Runze and Ma, Kai and Huang, Linjiang and Huang, Shaofei and Gao, Jialin and Wei, Xiaoming and Dai, Jiao and Han, Jizhong and Liu, Si},
  journal={arXiv preprint arXiv:2409.18071},
  year={2024}
}

@inproceedings{yu2025omnipaint,
  title={Omnipaint: Mastering object-oriented editing via disentangled insertion-removal inpainting},
  author={Yu, Yongsheng and Zeng, Ziyun and Zheng, Haitian and Luo, Jiebo},
  booktitle={Proceedings of the IEEE/CVF International Conference on Computer Vision},
  year={2025}
}

@inproceedings{zhang2025ultra,
  title={Ultra High-Resolution Image Inpainting with Patch-Based Content Consistency Adapter},
  author={Zhang, Jianhui and Cheng, Shen and Sun, Qirui and Liu, Jia and Luyang, Wang and Feng, Chaoyu and Fang, Chen and Lei, Lei and Wang, Jue and Liu, Shuaicheng},
  booktitle={Proceedings of the IEEE/CVF International Conference on Computer Vision},
  pages={16991--17000},
  year={2025}
}

@inproceedings{tang2025ata,
  title={ATA: Adaptive Transformation Agent for Text-Guided Subject-Position Variable Background Inpainting},
  author={Tang, Yizhe and Sun, Zhimin and Du, Yuzhen and Yi, Ran and Lu, Guangben and Hu, Teng and Li, Luying and Ma, Lizhuang and Zou, Fangyuan},
  booktitle={Proceedings of the Computer Vision and Pattern Recognition Conference},
  pages={18335--18345},
  year={2025}
}

@inproceedings{chen2025ctr,
  title={CTR-Driven Advertising Image Generation with Multimodal Large Language Models},
  author={Chen, Xingye and Feng, Wei and Du, Zhenbang and Wang, Weizhen and Chen, Yanyin and Wang, Haohan and Liu, Linkai and Li, Yaoyu and Zhao, Jinyuan and Li, Yu and others},
  booktitle={Proceedings of the ACM on Web Conference 2025},
  pages={2262--2275},
  year={2025}
}

@article{yi2024feditnet,
  title={Feditnet++: Few-shot editing of latent semantics in gan spaces with correlated attribute disentanglement},
  author={Yi, Ran and Hu, Teng and Xia, Mengfei and Tang, Yizhe and Liu, Yong-Jin},
  journal={IEEE Transactions on Pattern Analysis and Machine Intelligence},
  year={2024},
  publisher={IEEE}
}

@inproceedings{hu2024anomalydiffusion,
  title={Anomalydiffusion: Few-shot anomaly image generation with diffusion model},
  author={Hu, Teng and Zhang, Jiangning and Yi, Ran and Du, Yuzhen and Chen, Xu and Liu, Liang and Wang, Yabiao and Wang, Chengjie},
  booktitle={Proceedings of the AAAI conference on artificial intelligence},
  volume={38},
  number={8},
  pages={8526--8534},
  year={2024}
}

@inproceedings{hu2025improving,
  title={Improving autoregressive visual generation with cluster-oriented token prediction},
  author={Hu, Teng and Zhang, Jiangning and Yi, Ran and Weng, Jieyu and Wang, Yabiao and Zeng, Xianfang and Xue, Zhucun and Ma, Lizhuang},
  booktitle={Proceedings of the Computer Vision and Pattern Recognition Conference},
  pages={9351--9360},
  year={2025}
}

@article{yi2025iar2,
  title={IAR2: Improving Autoregressive Visual Generation with Semantic-Detail Associated Token Prediction},
  author={Yi, Ran and Hu, Teng and Su, Zihan and Ma, Lizhuang},
  journal={arXiv preprint arXiv:2510.06928},
  year={2025}
}

@article{bu2025hiflow,
  title={HiFlow: Training-free High-Resolution Image Generation with Flow-Aligned Guidance},
  author={Bu, Jiazi and Ling, Pengyang and Zhou, Yujie and Zhang, Pan and Wu, Tong and Dong, Xiaoyi and Zang, Yuhang and Cao, Yuhang and Lin, Dahua and Wang, Jiaqi},
  journal={arXiv preprint arXiv:2504.06232},
  year={2025}
}

@article{rafailov2023direct,
  title={Direct preference optimization: Your language model is secretly a reward model},
  author={Rafailov, Rafael and Sharma, Archit and Mitchell, Eric and Manning, Christopher D and Ermon, Stefano and Finn, Chelsea},
  journal={Advances in neural information processing systems},
  volume={36},
  pages={53728--53741},
  year={2023}
}

@article{christiano2017deep,
  title={Deep reinforcement learning from human preferences},
  author={Christiano, Paul F and Leike, Jan and Brown, Tom and Martic, Miljan and Legg, Shane and Amodei, Dario},
  journal={Advances in neural information processing systems},
  volume={30},
  year={2017}
}

@article{schulman2017proximal,
  title={Proximal policy optimization algorithms},
  author={Schulman, John and Wolski, Filip and Dhariwal, Prafulla and Radford, Alec and Klimov, Oleg},
  journal={arXiv preprint arXiv:1707.06347},
  year={2017}
}

@article{shao2024deepseekmath,
  title={Deepseekmath: Pushing the limits of mathematical reasoning in open language models},
  author={Shao, Zhihong and Wang, Peiyi and Zhu, Qihao and Xu, Runxin and Song, Junxiao and Bi, Xiao and Zhang, Haowei and Zhang, Mingchuan and Li, YK and Wu, Yang and others},
  journal={arXiv preprint arXiv:2402.03300},
  year={2024}
}

@article{ouyang2022training,
  title={Training language models to follow instructions with human feedback},
  author={Ouyang, Long and Wu, Jeffrey and Jiang, Xu and Almeida, Diogo and Wainwright, Carroll and Mishkin, Pamela and Zhang, Chong and Agarwal, Sandhini and Slama, Katarina and Ray, Alex and others},
  journal={Advances in neural information processing systems},
  volume={35},
  pages={27730--27744},
  year={2022}
}

@article{achiam2023gpt,
  title={Gpt-4 technical report},
  author={Achiam, Josh and Adler, Steven and Agarwal, Sandhini and Ahmad, Lama and Akkaya, Ilge and Aleman, Florencia Leoni and Almeida, Diogo and Altenschmidt, Janko and Altman, Sam and Anadkat, Shyamal and others},
  journal={arXiv preprint arXiv:2303.08774},
  year={2023}
}

@article{kaufmann2024survey,
  title={A survey of reinforcement learning from human feedback},
  author={Kaufmann, Timo and Weng, Paul and Bengs, Viktor and H{\"u}llermeier, Eyke},
  year={2024}
}

@article{xu2023imagereward,
  title={Imagereward: Learning and evaluating human preferences for text-to-image generation},
  author={Xu, Jiazheng and Liu, Xiao and Wu, Yuchen and Tong, Yuxuan and Li, Qinkai and Ding, Ming and Tang, Jie and Dong, Yuxiao},
  journal={Advances in Neural Information Processing Systems},
  volume={36},
  pages={15903--15935},
  year={2023}
}

@inproceedings{wallace2024diffusion,
  title={Diffusion model alignment using direct preference optimization},
  author={Wallace, Bram and Dang, Meihua and Rafailov, Rafael and Zhou, Linqi and Lou, Aaron and Purushwalkam, Senthil and Ermon, Stefano and Xiong, Caiming and Joty, Shafiq and Naik, Nikhil},
  booktitle={Proceedings of the IEEE/CVF Conference on Computer Vision and Pattern Recognition},
  pages={8228--8238},
  year={2024}
}

@inproceedings{black2023training,
  title={Training Diffusion Models with Reinforcement Learning},
  author={Black, Kevin and Janner, Michael and Du, Yilun and Kostrikov, Ilya and Levine, Sergey},
  booktitle={The Twelfth International Conference on Learning Representations}
}

@inproceedings{yang2024using,
  title={Using human feedback to fine-tune diffusion models without any reward model},
  author={Yang, Kai and Tao, Jian and Lyu, Jiafei and Ge, Chunjiang and Chen, Jiaxin and Shen, Weihan and Zhu, Xiaolong and Li, Xiu},
  booktitle={Proceedings of the IEEE/CVF Conference on Computer Vision and Pattern Recognition},
  pages={8941--8951},
  year={2024}
}

@article{fan2023dpok,
  title={Dpok: Reinforcement learning for fine-tuning text-to-image diffusion models},
  author={Fan, Ying and Watkins, Olivia and Du, Yuqing and Liu, Hao and Ryu, Moonkyung and Boutilier, Craig and Abbeel, Pieter and Ghavamzadeh, Mohammad and Lee, Kangwook and Lee, Kimin},
  journal={Advances in Neural Information Processing Systems},
  volume={36},
  pages={79858--79885},
  year={2023}
}

@article{ye2023ip,
  title={Ip-adapter: Text compatible image prompt adapter for text-to-image diffusion models},
  author={Ye, Hu and Zhang, Jun and Liu, Sibo and Han, Xiao and Yang, Wei},
  journal={arXiv preprint arXiv:2308.06721},
  year={2023}
}

@inproceedings{zhang2019bertscore,
  title={BERTScore: Evaluating Text Generation with BERT},
  author={Zhang, Tianyi and Kishore, Varsha and Wu, Felix and Weinberger, Kilian Q and Artzi, Yoav},
  booktitle={International Conference on Learning Representations}
}

@article{xue2025dancegrpo,
  title={DanceGRPO: Unleashing GRPO on Visual Generation},
  author={Xue, Zeyue and Wu, Jie and Gao, Yu and Kong, Fangyuan and Zhu, Lingting and Chen, Mengzhao and Liu, Zhiheng and Liu, Wei and Guo, Qiushan and Huang, Weilin and others},
  journal={arXiv preprint arXiv:2505.07818},
  year={2025}
}

@inproceedings{radford2021learning,
  title={Learning transferable visual models from natural language supervision},
  author={Radford, Alec and Kim, Jong Wook and Hallacy, Chris and Ramesh, Aditya and Goh, Gabriel and Agarwal, Sandhini and Sastry, Girish and Askell, Amanda and Mishkin, Pamela and Clark, Jack and others},
  booktitle={International conference on machine learning},
  pages={8748--8763},
  year={2021},
  organization={PmLR}
}

@inproceedings{liang2025aesthetic,
  title={Aesthetic post-training diffusion models from generic preferences with step-by-step preference optimization},
  author={Liang, Zhanhao and Yuan, Yuhui and Gu, Shuyang and Chen, Bohan and Hang, Tiankai and Cheng, Mingxi and Li, Ji and Zheng, Liang},
  booktitle={Proceedings of the Computer Vision and Pattern Recognition Conference},
  pages={13199--13208},
  year={2025}
}

@inproceedings{zhangitercomp,
  title={IterComp: Iterative Composition-Aware Feedback Learning from Model Gallery for Text-to-Image Generation},
  author={Zhang, Xinchen and Yang, Ling and Li, Guohao and Cai, YaQi and Tang, Yong and Yang, Yujiu and Wang, Mengdi and CUI, Bin and others},
  booktitle={The Thirteenth International Conference on Learning Representations}
}

@inproceedings{hu2025towards,
  title={Towards better alignment: Training diffusion models with reinforcement learning against sparse rewards},
  author={Hu, Zijing and Zhang, Fengda and Chen, Long and Kuang, Kun and Li, Jiahui and Gao, Kaifeng and Xiao, Jun and Wang, Xin and Zhu, Wenwu},
  booktitle={Proceedings of the Computer Vision and Pattern Recognition Conference},
  pages={23604--23614},
  year={2025}
}

@article{liu2025flow,
  title={Flow-grpo: Training flow matching models via online rl},
  author={Liu, Jie and Liu, Gongye and Liang, Jiajun and Li, Yangguang and Liu, Jiaheng and Wang, Xintao and Wan, Pengfei and Zhang, Di and Ouyang, Wanli},
  journal={arXiv preprint arXiv:2505.05470},
  year={2025}
}

@article{wu2023human,
  title={Human preference score v2: A solid benchmark for evaluating human preferences of text-to-image synthesis},
  author={Wu, Xiaoshi and Hao, Yiming and Sun, Keqiang and Chen, Yixiong and Zhu, Feng and Zhao, Rui and Li, Hongsheng},
  journal={arXiv preprint arXiv:2306.09341},
  year={2023}
}

@article{kirstain2023pick,
  title={Pick-a-pic: An open dataset of user preferences for text-to-image generation},
  author={Kirstain, Yuval and Polyak, Adam and Singer, Uriel and Matiana, Shahbuland and Penna, Joe and Levy, Omer},
  journal={Advances in neural information processing systems},
  volume={36},
  pages={36652--36663},
  year={2023}
}

@inproceedings{karthik2025scalable,
  title={Scalable ranked preference optimization for text-to-image generation},
  author={Karthik, Shyamgopal and Coskun, Huseyin and Akata, Zeynep and Tulyakov, Sergey and Ren, Jian and Kag, Anil},
  booktitle={Proceedings of the IEEE/CVF International Conference on Computer Vision},
  pages={18399--18410},
  year={2025}
}

@inproceedings{huang2025patchdpo,
  title={Patchdpo: Patch-level dpo for finetuning-free personalized image generation},
  author={Huang, Qihan and Chan, Long and Liu, Jinlong and He, Wanggui and Jiang, Hao and Song, Mingli and Song, Jie},
  booktitle={Proceedings of the Computer Vision and Pattern Recognition Conference},
  pages={18369--18378},
  year={2025}
}

@article{li2024aligning,
  title={Aligning diffusion models by optimizing human utility},
  author={Li, Shufan and Kallidromitis, Konstantinos and Gokul, Akash and Kato, Yusuke and Kozuka, Kazuki},
  journal={Advances in Neural Information Processing Systems},
  volume={37},
  pages={24897--24925},
  year={2024}
}

@inproceedings{clark2023directly,
  title={Directly Fine-Tuning Diffusion Models on Differentiable Rewards},
  author={Clark, Kevin and Vicol, Paul and Swersky, Kevin and Fleet, David J},
  booktitle={The Twelfth International Conference on Learning Representations},
  year={2023}
}

@article{zheng2024bilateral,
  title={Bilateral Reference for High-Resolution Dichotomous Image Segmentation},
  author={Zheng, Peng and Gao, Dehong and Fan, Deng-Ping and Liu, Li and Laaksonen, Jorma and Ouyang, Wanli and Sebe, Nicu},
  journal={arXiv preprint arXiv:2401.03407},
  year={2024}
}

@article{hu2022lora,
  title={Lora: Low-rank adaptation of large language models.},
  author={Hu, Edward J and Shen, Yelong and Wallis, Phillip and Allen-Zhu, Zeyuan and Li, Yuanzhi and Wang, Shean and Wang, Lu and Chen, Weizhu and others},
  journal={ICLR},
  volume={1},
  number={2},
  pages={3},
  year={2022}
}

@article{tian2024hydralora,
  title={Hydralora: An asymmetric lora architecture for efficient fine-tuning},
  author={Tian, Chunlin and Shi, Zhan and Guo, Zhijiang and Li, Li and Xu, Cheng-Zhong},
  journal={Advances in Neural Information Processing Systems},
  volume={37},
  pages={9565--9584},
  year={2024}
}

@article{heusel2017gans,
  title={Gans trained by a two time-scale update rule converge to a local nash equilibrium},
  author={Heusel, Martin and Ramsauer, Hubert and Unterthiner, Thomas and Nessler, Bernhard and Hochreiter, Sepp},
  journal={Advances in neural information processing systems},
  volume={30},
  year={2017}
}

@inproceedings{lin2014microsoft,
  title={Microsoft coco: Common objects in context},
  author={Lin, Tsung-Yi and Maire, Michael and Belongie, Serge and Hays, James and Perona, Pietro and Ramanan, Deva and Doll{\'a}r, Piotr and Zitnick, C Lawrence},
  booktitle={Computer vision--ECCV 2014: 13th European conference, zurich, Switzerland, September 6-12, 2014, proceedings, part v 13},
  pages={740--755},
  year={2014},
  organization={Springer}
}

@article{ravi2024sam2,
  title={SAM 2: Segment Anything in Images and Videos},
  author={Ravi, Nikhila and Gabeur, Valentin and Hu, Yuan-Ting and Hu, Ronghang and Ryali, Chaitanya and Ma, Tengyu and Khedr, Haitham and R{\"a}dle, Roman and Rolland, Chloe and Gustafson, Laura and Mintun, Eric and Pan, Junting and Alwala, Kalyan Vasudev and Carion, Nicolas and Wu, Chao-Yuan and Girshick, Ross and Doll{\'a}r, Piotr and Feichtenhofer, Christoph},
  journal={arXiv preprint arXiv:2408.00714},
  url={https://arxiv.org/abs/2408.00714},
  year={2024}
}

@inproceedings{lin2025evaluating,
  title={Evaluating text-to-visual generation with image-to-text generation},
  author={Lin, Zhiqiu and Pathak, Deepak and Li, Baiqi and Li, Jiayao and Xia, Xide and Neubig, Graham and Zhang, Pengchuan and Ramanan, Deva},
  booktitle={European Conference on Computer Vision},
  pages={366--384},
  year={2025},
  organization={Springer}
}

@article{chen2024virtualmodel,
  title={VirtualModel: Generating Object-ID-retentive Human-object Interaction Image by Diffusion Model for E-commerce Marketing},
  author={Chen, Binghui and Zhong, Chongyang and Xiang, Wangmeng and Geng, Yifeng and Xie, Xuansong},
  journal={arXiv preprint arXiv:2405.09985},
  year={2024}
}

@article{oquab2023dinov2,
  title={Dinov2: Learning robust visual features without supervision},
  author={Oquab, Maxime and Darcet, Timoth{\'e}e and Moutakanni, Th{\'e}o and Vo, Huy and Szafraniec, Marc and Khalidov, Vasil and Fernandez, Pierre and Haziza, Daniel and Massa, Francisco and El-Nouby, Alaaeldin and others},
  journal={arXiv preprint arXiv:2304.07193},
  year={2023}
}

@article{elo1978rating,
  title={The rating of chessplayers: Past and present},
  author={Elo, Arpad E and Sloan, Sam},
  journal={Ishi Press},
  year={1978}
}

@article{labs2025fluxkontext,
  title={FLUX. 1 Kontext: Flow Matching for In-Context Image Generation and Editing in Latent Space},
  author={Labs, Black Forest and Batifol, Stephen and Blattmann, Andreas and Boesel, Frederic and Consul, Saksham and Diagne, Cyril and Dockhorn, Tim and English, Jack and English, Zion and Esser, Patrick and others},
  journal={arXiv preprint arXiv:2506.15742},
  year={2025}
}

@article{wu2025qwen,
  title={Qwen-image technical report},
  author={Wu, Chenfei and Li, Jiahao and Zhou, Jingren and Lin, Junyang and Gao, Kaiyuan and Yan, Kun and Yin, Sheng-ming and Bai, Shuai and Xu, Xiao and Chen, Yilei and others},
  journal={arXiv preprint arXiv:2508.02324},
  year={2025}
}

@inproceedings{xiao2024florence,
  title={Florence-2: Advancing a unified representation for a variety of vision tasks},
  author={Xiao, Bin and Wu, Haiping and Xu, Weijian and Dai, Xiyang and Hu, Houdong and Lu, Yumao and Zeng, Michael and Liu, Ce and Yuan, Lu},
  booktitle={Proceedings of the IEEE/CVF Conference on Computer Vision and Pattern Recognition},
  pages={4818--4829},
  year={2024}
}

@article{loshchilov2017decoupled,
  title={Decoupled weight decay regularization},
  author={Loshchilov, Ilya and Hutter, Frank},
  journal={arXiv preprint arXiv:1711.05101},
  year={2017}
}

@article{baldridge2024imagen,
  title={Imagen 3},
  author={Baldridge, Jason and Bauer, Jakob and Bhutani, Mukul and Brichtova, Nicole and Bunner, Andrew and Chan, Kelvin and Chen, Yichang and Dieleman, Sander and Du, Yuqing and Eaton-Rosen, Zach and others},
  journal={arXiv preprint arXiv:2408.07009},
  year={2024}
}

@misc{diffsynth,
    author={ModelScope},
    title={DiffSynth-Studio},
    year={2025},
    howpublished={\url{https://github.com/modelscope/DiffSynth-Studio}},
}

@article{ho2020denoising,
  title={Denoising diffusion probabilistic models},
  author={Ho, Jonathan and Jain, Ajay and Abbeel, Pieter},
  journal={Advances in neural information processing systems},
  volume={33},
  pages={6840--6851},
  year={2020}
}

@inproceedings{sohl2015deep,
  title={Deep unsupervised learning using nonequilibrium thermodynamics},
  author={Sohl-Dickstein, Jascha and Weiss, Eric and Maheswaranathan, Niru and Ganguli, Surya},
  booktitle={International conference on machine learning},
  pages={2256--2265},
  year={2015},
  organization={pmlr}
}
}
\clearpage
\appendix
\section*{\LARGE Appendix}
\section{Overview}

In this supplementary material, we mainly present the following contents:
\begin{itemize}
    \item More implementation details of our model structure (Sec.~\ref{implementation}). 
    \item Detailed derivation analysis of Shared Commonality Preference Optimization (Sec.~\ref{derivation}).
    \item More technical details of the evaluation metrics (Sec.~\ref{metrics}).
    \item More details of the user study (Sec.~\ref{userstudy}).
    \item More qualitative comparisons with state-of-the-art inpainting and editing methods (Sec.~\ref{more_qualitative}).
    \item More ablation study on the Conditional Asymmetric Preference Optimization (Sec.~\ref{ablaiton_capo}).
    \item More experiments settings and results of method migration on Qwen-Image-Edit (Sec.~\ref{migration}).
    \item More results of InpaintDPO with different aspect ratios (Sec.~\ref{more_results}).
\end{itemize}

\section{Implementation Details}
\label{implementation}
\begin{figure*}[t]
    \centering
    \includegraphics[width=0.85\linewidth]{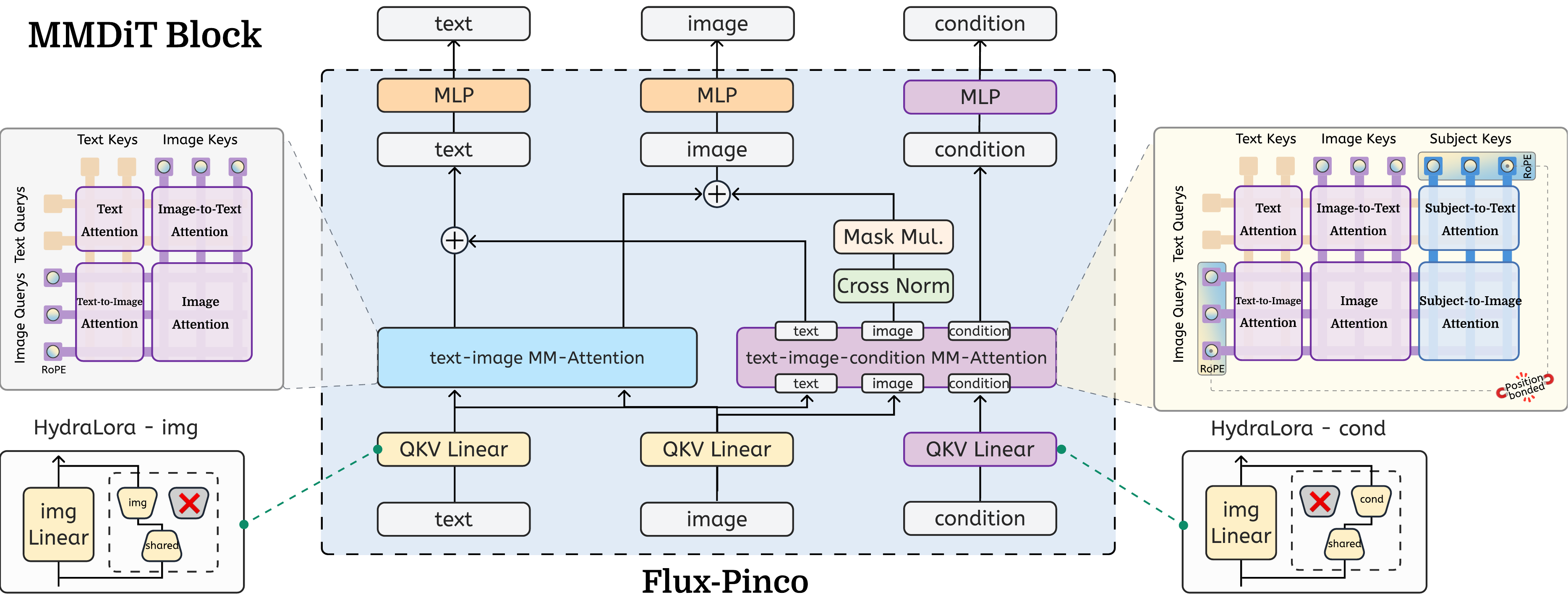}
    \vspace{-0.1in}
    \caption{Architecture of Flux-Pinco, the base model of our InpaintDPO.}
    \label{fig: flux-pinco}
\end{figure*}

\textbf{Backbone}. We apply our proposed InpaintDPO on Flux-Pinco~\cite{lu2025pinco}. 
We use the VAE Encoder of the original Flux model as our semantic feature extractor and only take the output of the final layer as our semantic feature. 
Meanwhile, we also construct a simple convolutional network (ConvNet) to extract the shape feature from the foreground mask.
The architecture of the MMDiT block in Flux-Pinco is illustrated in Fig. \ref{fig: flux-pinco}. 
Specifically, in addition to the conventional text-image MM-Attention block, Pinco introduces a text-image-condition MM-Attention mechanism. 
This design links the subject description in the text and image with the subject encoded in the condition, thereby enabling the model to correctly associate the subject in the text with the subject image provided by the condition, which helps to mitigate the issue of subject expansion. 
Furthermore, considering that both image and condition operations are performed at the image level, we employ HydraLoRA \cite{tian2024hydralora} to reduce the number of trainable parameters. 
HydraLoRA allows for efficient training by sharing the majority of parameters between the LoRA modules for the image and condition branches.

\section{Derivation of SCPO}
\label{derivation}
In Sec.3.4 in the main paper, we rewrite the Bradley-Terry (BT) model to learn the shared commonality preference between win-win pairs:
\begin{equation}
\label{eq:bt}
    p_{\text{BT}}(\boldsymbol{x}_0^{w_1}\sim\boldsymbol{x}_0^{w_2} | \boldsymbol{c}) = 1-\sigma(\left| r(\boldsymbol{c}, \boldsymbol{x}_0^{w_1}) - r(\boldsymbol{c}, \boldsymbol{x}_0^{w_2}) \right| ).
\end{equation}
RLHF aims to optimize a conditional distribution $p_\theta(x_0|c)$ (conditioning $c \sim \mathcal{D}_c$) such that the latent reward model $r(c, x_0)$ defined on it is maximized, while regularizing the KL-divergence from a reference distribution $p_{\text{ref}}$:
\begin{align}
\label{eq:kl}
\max_{p_\theta} &\ \mathbb{E}_{c \sim \mathcal{D}_c, x_0 \sim p_\theta(x_0|c)} \left| r(c, x_0) \right| \notag \\
&- \beta \mathbb{D}_{\text{KL}} \left| p_\theta(x_0|c) \| p_{\text{ref}}(x_0|c) \right|.
\end{align}
From \cite{rafailov2023direct}, the unique global optimal solution \( p_\theta^* \) becomes:
\begin{equation}
p_\theta^*(x_0|c) = p_{\text{ref}}(x_0|c) \exp\left(r(c, x_0)/\beta\right)/Z(c),
\end{equation}
where \( Z(c) = \sum_{x_0} p_{\text{ref}}(x_0|c) \exp\left(r(c, x_0)/\beta\right) \) is the partition function. Hence, the reward function is rewritten as:
\begin{equation}
r(c, x_0) = \beta \log \frac{p_\theta^*(x_0|c)}{p_{\text{ref}}(x_0|c)} + \beta \log Z(c).
\end{equation}
To adapt SCPO to diffusion models, we introduce latents \( x_{1:T} \) and define \( R(c, x_{0:T}) \) as the reward on the whole chain, such that we can define \( r(c, x_0) \) as:
\begin{equation}
r(c, x_0) = \mathbb{E}_{p_\theta(x_{1:T}|x_0, c)} \left| R(c, x_{0:T}) \right|.
\end{equation}
As for the KL-regularization term in Eq.~\ref{eq:kl} , following prior works \cite{ho2020denoising,sohl2015deep}, we can instead minimize KL-regularization term's upper bound joint KL-divergence.
So we have the objective:
\begin{align}
\max_{p_\theta} &\ \mathbb{E}_{c \sim \mathcal{D}_c, x_{0:T} \sim p_\theta(x_{0:T}|c)} \left| r(c, x_0) \right| \notag \\
&- \beta \mathbb{D}_{\text{KL}} \left| p_\theta(x_{0:T}|c) \| p_{\text{ref}}(x_{0:T}|c) \right|.
\end{align}
With Eq. \ref{eq:bt}, the reward objective becomes:
\begin{align}
&L_{\text{SCPO}}(\theta) = -\mathbb{E}_{(x_0^{w_1}, x_0^{w_2}) \sim \mathcal{D}} \log \Bigg( 1-\sigma\bigg( \notag \\
&\quad \beta \mathbb{E}_{\substack{x_{1:T}^{w_1} \sim p_\theta(x_{1:T}^{w_1}|x_0^{w_1}) \\ x_{1:T}^{w_2} \sim p_\theta(x_{1:T}^{w_2}|x_0^{w_2})}} \Biggl| \log \frac{p_\theta(x_{0:T}^{w_1})}{p_{\text{ref}}(x_{0:T}^{w_1})} - \log \frac{p_\theta(x_{0:T}^{x_2})}{p_{\text{ref}}(x_{0:T}^{x_2})} \Biggr| \bigg) \Bigg).
\end{align}
Since sampling from $p_\theta(x_{1:T}|x_0^{w_1})$ is intractable, we utilize $q(x_{1:T}|x_0^{w_2})$ for approximation:
\begin{align}
L_1(\theta) &= -\log \Bigg( 1-\sigma\bigg( \beta \mathbb{E}_{\substack{x_{1:T}^{w_1} \sim q(x_{1:T}|x_0^{w_1}) \\ x_{1:T}^{w_2} \sim q(x_{1:T}|x_0^{w_2})}} \notag \\
&\quad \Biggl| \log \frac{p_\theta(x_{0:T}^{w_1})}{p_{\text{ref}}(x_{0:T}^{w_1})} - \log \frac{p_\theta(x_{0:T}^{x_2})}{p_{\text{ref}}(x_{0:T}^{x_2})} \Biggr| \bigg) \Bigg) \notag \\
&= -\log \Bigg( 1-\sigma\bigg( \beta \mathbb{E}_{\substack{x_{1:T}^{w_1} \sim q(x_{1:T}|x_0^{w_1}) \\ x_{1:T}^{w_2} \sim q(x_{1:T}|x_0^{w_2})}} \notag \\
&\quad \Biggl| \sum_{t=1}^T \log \frac{p_\theta(x_{t-1}^{w_1}|x_t^{w_1})}{p_{\text{ref}}(x_{t-1}^{w_1}|x_t^{w_1})} - \log \frac{p_\theta(x_{t-1}^{w_2}|x_t^{w_2})}{p_{\text{ref}}(x_{t-1}^{w_2}|x_t^{w_2})} \Biggr| \bigg) \Bigg).
\end{align}
By Triangle inequality, we have:
\begin{align}
L_1(\theta) &\leq -\log \Bigg( 1-\sigma\bigg( \beta \mathbb{E}_{\substack{x_{1:T}^{w_1} \sim q(x_{1:T}|x_0^{w_1}) \\ x_{1:T}^{w_2} \sim q(x_{1:T}|x_0^{w_2})}}  \notag \\
&\quad T \mathbb{E}_t \Biggl| \log \frac{p_\theta(x_{t-1}^{w_1}|x_t^{w_1})}{p_{\text{ref}}(x_{t-1}^{w_1}|x_t^{w_1})} - \log \frac{p_\theta(x_{t-1}^{w_2}|x_t^{w_2})}{p_{\text{ref}}(x_{t-1}^{w_2}|x_t^{w_2})} \Biggr| \bigg) \Bigg) \notag \\
&=-\log \Bigg( 1-\sigma\bigg( \beta T \mathbb{E}_t \mathbb{E}_{\substack{x_{t-1,t}^{w_1} \sim q(x_{t-1,t}|x_0^{w_1}) \\ x_{t-1,t}^{w_2} \sim q(x_{t-1,t}|x_0^{w_2})}}  \notag \\
&\quad \Biggl| \log \frac{p_\theta(x_{t-1}^{w_1}|x_t^{w_1})}{p_{\text{ref}}(x_{t-1}^{w_1}|x_t^{w_1})} - \log \frac{p_\theta(x_{t-1}^{w_2}|x_t^{w_2})}{p_{\text{ref}}(x_{t-1}^{w_2}|x_t^{w_2})} \Biggr| \bigg) \Bigg).
\end{align}
With some algebra and applying Jensen's inequality, we have:
\begin{align}
L(\theta) = &-\mathbb{E}_{\substack{(x_0^{w_1}, x_0^{w_2}) \sim \mathcal{D}, t \sim \mathcal{U}(0,T) \\ x_t^{w_1} \sim q(x_t^{w_1}|x_0^{w_1}) \\ x_t^{w_2} \sim q(x_t^{w_2}|x_0^{w_2})}} \log \Bigg( 1- \sigma\bigg(\beta T \Biggl| \notag \\
&\quad \mathbb{D}_{\text{KL}}(q(x_{t-1}^{w_1}|x_{0},x_{t}^{w_1}) \| p_\theta(x_{t-1}^{w_1}|x_t^{w_1})) \notag \\
&\quad - \mathbb{D}_{\text{KL}}(q(x_{t-1}^{w_1}|x_{0},x_{t}^{w_1}) \| p_{\text{ref}}(x_{t-1}^{w_1}|x_t^{w_1})) \notag \\
&\quad - \mathbb{D}_{\text{KL}}(q(x_{t-1}^{w_2}|x_{0},x_{t}^{w_2}) \| p_\theta(x_{t-1}^{w_2}|x_t^{w_2})) \notag \\
&\quad + \mathbb{D}_{\text{KL}}(q(x_{t-1}^{w_2}|x_{0},x_{t}^{w_2}) \| p_{\text{ref}}(x_{t-1}^{w_2}|x_t^{w_2})) \Biggr|\bigg)\Bigg).
\end{align}
Then, we can yield the final SCPO loss:
\begin{equation}
\begin{split}
&L_{\text{SCPO}}(\theta) = -\mathbb{E} \log \Bigg(1 - \sigma \bigg( \beta T \omega (\lambda_t) \Biggl| \notag \\
&\quad \left( \|\bm{\epsilon}^{w_1} - \bm{\epsilon}_\theta(\bm{x}_t^{w_1}, t)\|_2^2 - \|\bm{\epsilon}^{w_1} - \bm{\epsilon}_{\text{ref}}(\bm{x}_t^{w_1}, t)\|_2^2 \right) \notag \\
&\quad - \left( \|\bm{\epsilon}^{w_2} - \bm{\epsilon}_\theta(\bm{x}_t^{w_2}, t)\|_2^2 - \|\bm{\epsilon}^{w_2} - \bm{\epsilon}_{\text{ref}}(\bm{x}_t^{w_2}, t)\|_2^2 \right) \Biggr| \bigg) \Bigg).
\end{split}
\end{equation}

\begin{figure*}[t]
    \centering
    \includegraphics[width=0.8\linewidth]{figs/GPT4O_QA-2.pdf}
    % \vspace{-0.1in}
    \caption{GPT-4o prompt for assessment and its reply. Note that you need to specify the name of the subject in \textcolor[RGB]{14,0,255}{[subject]}.}
    \label{fig:gpt4o qa}
\end{figure*}

\section{Evaluation Metrics}
\label{metrics}
In this section, we provide a more detailed description of the evaluation metrics.
We quantitatively evaluate the model's performance from the following four perspectives: Foreground Consistency, Spatial Rationality, Text-alignment, and Image Quality.

(1) \textbf{Foreground Consistency}: To evaluate the subject extension ratio, we employ the Object Extension Ratio (OER)~\cite{chen2024virtualmodel} metric, which quantifies the spatial consistency between the foreground subject mask of the generated image and the corresponding ground truth subject mask.
Specifically, we utilize BiRefNet~\cite{zheng2024bilateral} and SAM2.1~\cite{ravi2024sam2} respectively to segment the generated image, yielding an accurate foreground subject mask $M$.
For foreground-conditioned inpainting where the subject's position is fixed, the subject mask of the original image, denoted as $M_o$, serves as the ground truth mask (where $M_o$ is typically derived from the input mask $m$, i.e., $M_o = 1 - m$).
The OER score is then computed as follows:
\begin{equation}
    \text{OER} = \frac{\sum \text{ReLU}(M - M_o)}{\sum M_o},
\end{equation}
where $\text{ReLU}(\cdot)$ is the Rectified Linear Unit activation function.
The OER metric quantifies the extension of the subject in the generated image relative to the ground truth. 
Therefore, a smaller OER score indicates superior subject consistency achieved by the inpainting model.

(2) \textbf{Spatial Rationality}: We leverage the GPT4o~\cite{achiam2023gpt} to evaluate each image based on Spatial Scale, Positional Relationship, and Viewpoint.
The assessing standard for these three aspects and rating criteria are as follows:
\begin{itemize}
    \item Spatial Scale: Assess whether the size proportions between the subject and other objects in the image are realistic and whether there is any disproportion between the subject and surrounding objects.
    \item Positional relationship: Check whether the spatial relationship between the subject and other objects in the image is reasonable and consistent with common placement methods in daily life. Determine if the subject is placed in a physically impossible position, such as floating.
    \item Viewpoint: Examine whether the perspective extension direction of the product is consistent with that of its supporting surface. Judge if there is an unreasonable perspective relationship, such as mismatched horizontal lines or inconsistent convergence trends between the product and the supporting surface, which violates the basic principles of linear perspective.
    \item Rating Criteria: 1 point: Obvious errors, inconsistent with the real world. 3 points: Minor errors, somewhat inconsistent with the real world. 5 points: No obvious errors, consistent with the real world.
\end{itemize}
The detailed prompt given to GPT-4o and the reply are shown in Fig.~\ref{fig:gpt4o qa}.
We also show some rationality analysis returned by GPT-4o under the criteria mentioned above, as shown in Fig.~\ref{fig: gpt4o}.

\begin{figure}[t]
    \centering
    \vspace{-0.1in}
    \includegraphics[width=0.9\linewidth]{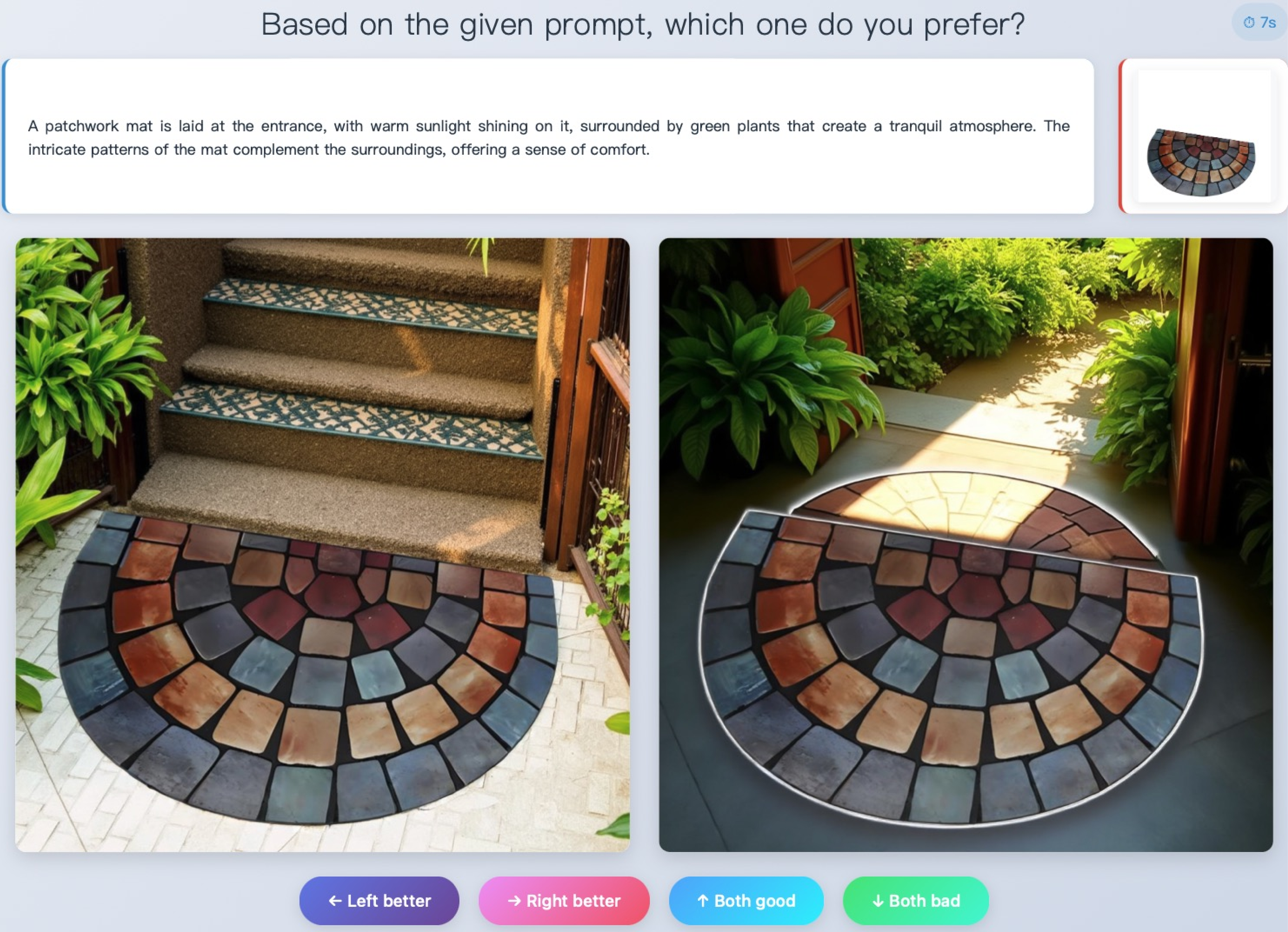}
    % \vspace{-0.1in}
    \caption{GUI of AI Inpainting Arena.}
    \label{fig: userstudy}
    % \vspace{-0.2in}
\end{figure}

(3) \textbf{Text-alignment}:
We evaluate text alignment from two aspects:
\begin{itemize}
    \item To measure the text-image alignment, we follow Imagen3 \cite{baldridge2024imagen} and adopt the Visual Question Answering (VQA) score \cite{lin2025evaluating}.
    This metric assesses the alignment between the generated image and the input text prompt by leveraging a VQA model to answer a series of simple yes-or-no questions regarding the image content.
    \item In the text-guided foreground-conditioned inpainting task, a common failure mode arises when the model redundantly generates the given foreground subject, often due to an inability to recognize the subject as already present based on the prompt.
    This often leads to generated results containing multiple similar subjects, which typically deviates from the user's intent.
    To quantitatively assess this failure, we utilize FlorenceV2 \cite{xiao2024florence} to detect and count the number of the specified subject within the generated image given the subject's name.
    Subsequently, the Multi-subject Rate is calculated as the ratio where the detected subject count exceeds the desired number of subjects. 
    A higher Multi-subject Rate indicates poorer subject recognition and control.
\end{itemize}

(4) \textbf{Image Quality}:
We evaluate the image quality from two aspects:
\begin{itemize}
    \item To evaluate the perceptual quality of the generated images, we compute the Fréchet Inception Distance (FID) \cite{heusel2017gans} score on the MSCOCO dataset \cite{lin2014microsoft}.
    \item Following OmniPaint~\cite{yu2025omnipaint}, we calculate the Context Coherence to quantify the structural consistency, since sometimes the foreground subject might exhibit structural misalignment with the surrounding background in the generated results.
    Specifically, denoting the foreground region as $\Omega_M$ and the background bounding box region excluding the foreground region as $\Omega_{B \setminus M}$, we compute the feature deviation by:
    \begin{equation}
        d_{\text{context}} = 1 - f(\Omega_M)^\top f(\Omega_{B \setminus M}),
    \end{equation}
    where the feature embeddings $f(\Omega)$ are extracted using the pre-trained vision model DINOv2~\cite{oquab2023dinov2}.

\end{itemize}

\begin{table}[t]
\centering
% \vspace{-0.2in}
\caption{Quantitative ablation studies on foreground-background coherence in CAPO.
}
\label{tab:capo}
\vspace{-0.1in}
\setlength\tabcolsep{2pt}
\scalebox{0.45}{
\begin{tabular}{ll|cc|c|c|cc}
\toprule
\multicolumn{2}{c|}{\multirow{2}{*}{\textbf{Combination}}} & \multicolumn{2}{c|}{\textbf{Foreground Consistency}} & \textbf{Spatial Rationality} & \textbf{Text-alignment} & \multicolumn{2}{c}{\textbf{Image Quality}} \\
\multicolumn{2}{c|}{} & \textbf{OER(BiRefNet)} $\downarrow$ & \textbf{OER(SAM2.1)}$\downarrow$ & \textbf{Avg.}$\uparrow$ & \textbf{VQA Score} $\uparrow$ & {\textbf{FID} $\downarrow$} & \textbf{Context}$\uparrow$ \\
\midrule
\multicolumn{2}{c|}{\textbf{MaskDPO}} & 5.6\% & 3.4\% & 3.93 & 0.863 & 98.87 & 0.250 \\
\multicolumn{2}{c|}{\textbf{MaskDPO-Expanded}} & 5.8\% & 3.4\% & 3.87 & 0.862 & 99.31 & 0.252 \\
\multicolumn{2}{c|}{\textbf{MaskDPO+CAPO}} & \textbf{5.6\%} & \textbf{3.1\%} & \textbf{3.93} & \textbf{0.866} & \textbf{97.42} & \textbf{0.258} \\
\bottomrule
\end{tabular}
}
\vspace{-0.1in}
\end{table}

\begin{figure}[t]
    \centering
    \includegraphics[width=\linewidth]{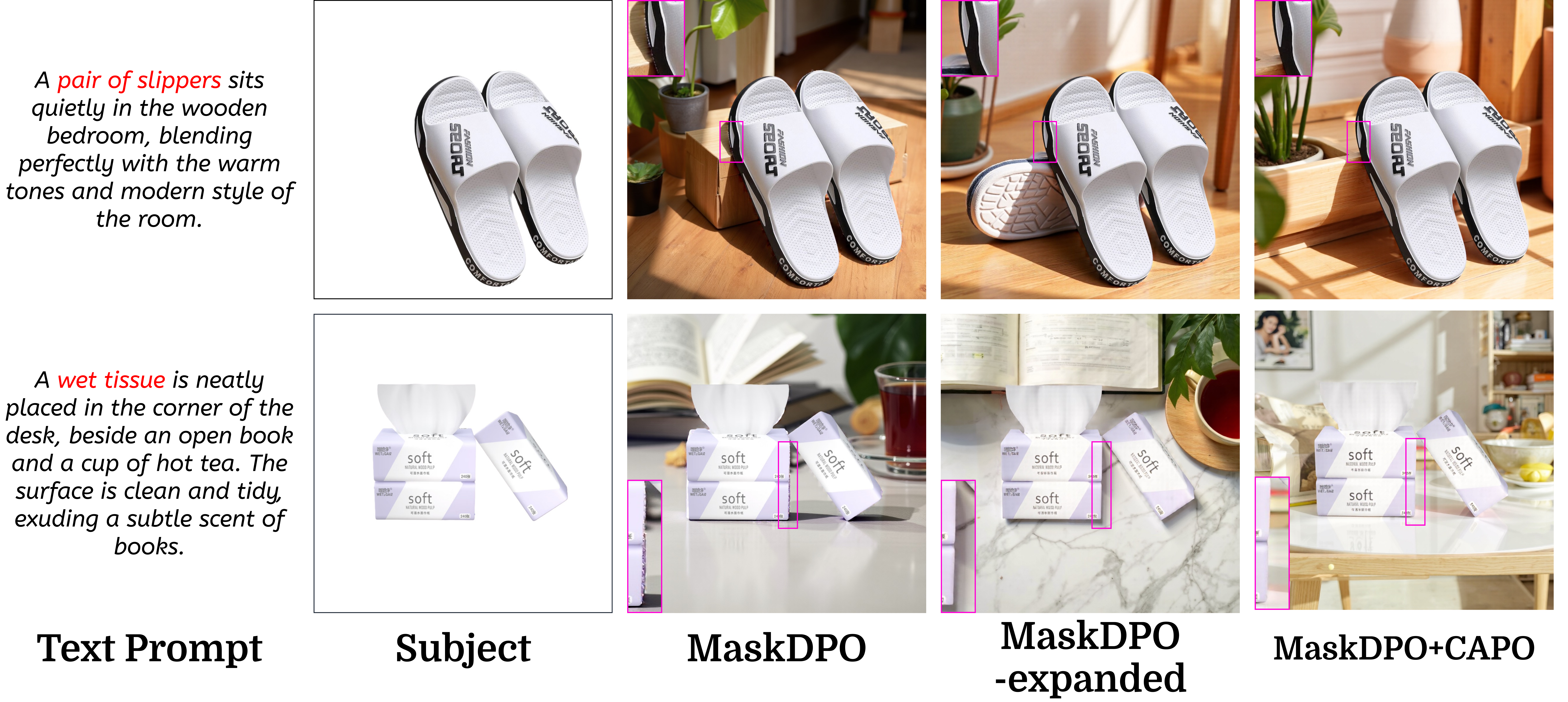}
    \vspace{-0.1in}
    \caption{\yr{Qualitative ablation study on CAPO.} %between MaskDPO, MaskDPO-Expanded and our MaskDPO+CAPO.
    }
    \label{fig: boundary_supp}
\end{figure}

\section{User Study}
\label{userstudy}
To comprehensively evaluate the perceived quality and, crucially, the spatial rationality of the generated images, we conducted a large-scale human preference study using an AI Inpainting Arena platform. 
This setup employs the ELO rating system \cite{elo1978rating}, a robust method traditionally used in competitive ranking, to accurately measure human preference for different inpainting methods.
Specifically, users were presented with a pair of anonymously generated images, each produced by a different model but sharing the identical input: the foreground image and the text prompt. 
The users' task was to act as judges and select the better image between the two, as shown in Fig.~\ref{fig: userstudy}. 
The primary criterion for the judgment was Spatial Rationality: the natural and visually consistent integration of the foreground subject within the generated background scene. 
This involves consideration of factors such as relative scale, spatial relationship, and viewpoint consistency between the foreground and the generated background.

We involved all the inpainting methods, including PowerPaint~\cite{zhuang2023powerpaint}, BrushNet~\cite{ju2024brushnet}, SD3-ControlNet~\cite{alimama_SD3}, OminiControl-inpaint~\cite{tan2025ominicontrol}, Flux-Fill-dev~\cite{flux2024inpaint}, Flux-ControlNet~\cite{alimama_FLUX}, Flux-Pinco~\cite{lu2025pinco}, and our InpaintDPO under comparison in this arena. 
To ensure high-quality and reliable judgments, we invited over 30 experts specialized in computer vision and AIGC fields to participate. 
After the collection of over 10,000 votes across numerous model competitions, the final ELO scores are calculated and presented in Tab.~2 in main paper, where our InpaintDPO model ranks first with the highest ELO score. 
These results strongly validate that our method is significantly preferred by human, particularly regarding the crucial aspect of spatial rationality.

\section{More Qualitative Comparisons}
\label{more_qualitative}
We provide more qualitative comparisons between our InpaintDPO with the state-of-the-art methods in Fig.~\ref{fig:more vision 1}, ~\ref{fig:more vision 2}, and ~\ref{fig:more vision 3}.
The compared baselines include:
\begin{itemize}
    \item Editing Methods: Flux-Kontext~\cite{labs2025fluxkontext} and Qwen-Image-Edit~\cite{wu2025qwen};
    \item UNet-based Inpainting Methods: PowerPaint~\cite{zhuang2023powerpaint} and BrushNet~\cite{ju2024brushnet};
    \item MMDiT-based Inpainting Methods: SD3-ControlNet~\cite{alimama_SD3}, OminiControl-inpaint~\cite{tan2025ominicontrol}, Flux-Fill-dev~\cite{flux2024inpaint}, Flux-ControlNet~\cite{alimama_FLUX} and Flux-Pinco~\cite{lu2025pinco}.
\end{itemize}

\yr{Results show our InpaintDPO consistently achieves better background spatial relationship rationality, while also maintaining superior performance in foreground consistency and text alignment; while current SOTA methods suffer from spatial relationship hallucinations, e.g., incorrect layout arrangements and object interaction, or worse foreground subject preservation.}

\section{More Ablation Study}
\label{ablaiton_capo}
\textbf{Boundary Artifacts Reduction}. 
In the main paper, in Sec. 4.3, we introduce CAPO, which improves boundary coherence by guiding the model with a global preference loss. 
To verify the effectiveness of our method, we introduce another naive approach for comparison: applying the erosion operation to the background mask. This process is intended to expand the effective scope of the inpainting loss in the subject area, thus creating a transitional area outside of the foreground subject for two distinct optimization losses. 
The comparison results are shown in Fig.~\ref{fig: boundary_supp} and Tab.~\ref{tab:capo}. Although the erosion can mitigate boundary artifacts, it ignores preference optimization in the transitional area, which includes foreground-background interaction information. Thus, it will to some extent affect the learning of positional rationality. Conversely, our CAPO not only maintains the integrity of MaskDPO and achieves superior boundary coherence, but also learns better positional rationality, thereby delivering superior performance in image quality and spatial rationality.

\section{More Results of Method Migration}
\label{migration}
\yr{We apply our InpaintDPO to Qwen-Image-Edit to enhance its spatial rationality. Specifically,}
we finetuned the Qwen-Image-Edit model with LoRA based on the repo~\cite{diffsynth}.
Since edit models often lack strong subject-following capability, we have increased the weight of the 
Inpainting Loss, where the hyperparameters $\lambda,\gamma,\mu$ are set to 5, 1, and 0.5, respectively. 
More results are shown in Fig. \ref{fig:qwen suppl}.
\yr{Results show applying our InpaintDPO successfully enabled the image editing model to achieve better subject detail preservation and improved spatial rationality, validating our InpaintDPO's generalizability.}

\section{More Results of Different Aspect Ratios}
\label{more_results}
\begin{figure*}[t]
    \centering
    \includegraphics[width=0.9\linewidth]{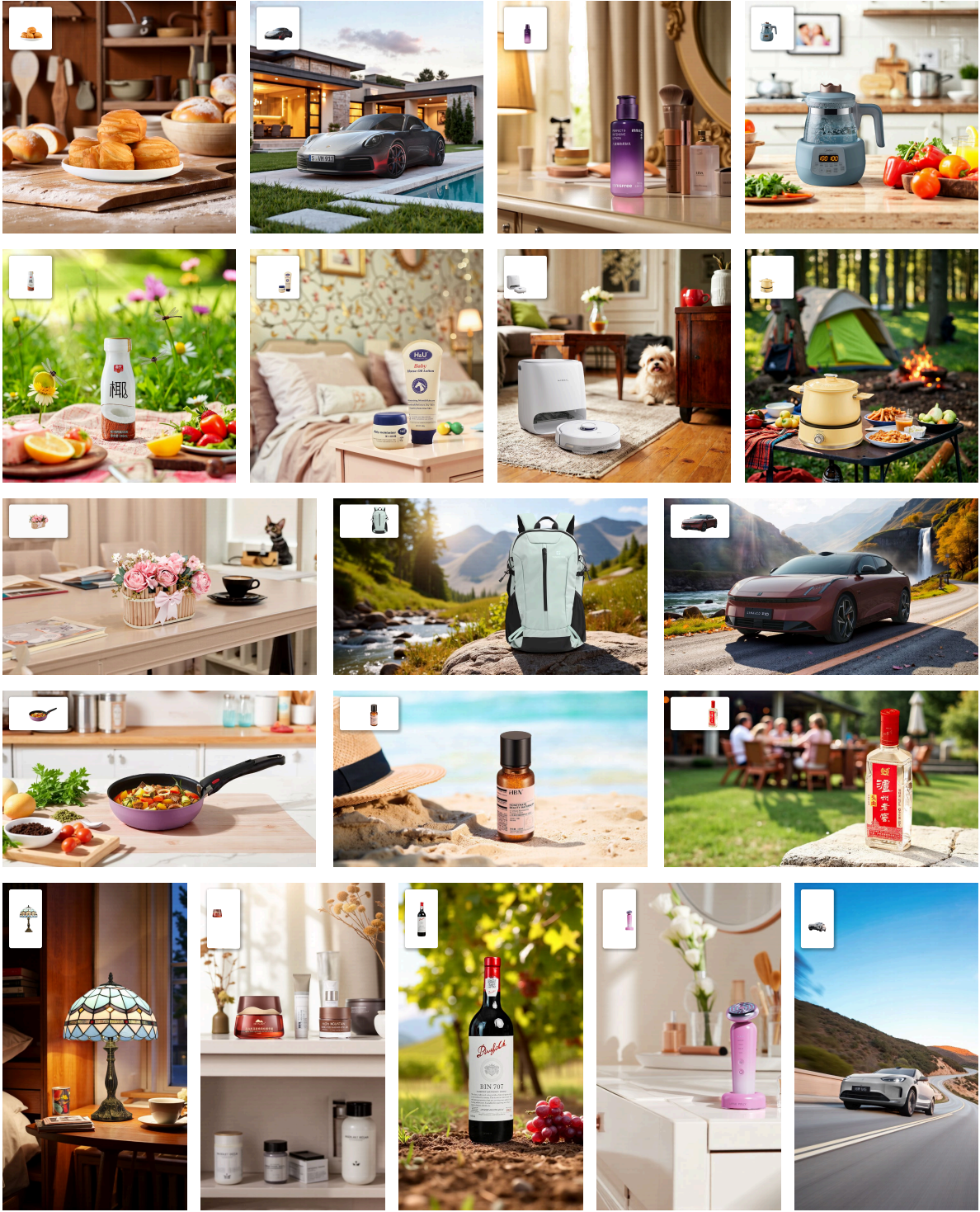}
    \vspace{-0.1in}
    \caption{More cases generated by InpaintDPO. InpaintDPO supports the generation of high-quality images with different aspect ratios, while ensuring the reasonable placement of subjects, achieving realistic foregroundconditioned inpainting.}
    \label{fig: morevision}
\end{figure*}

We show more generation results with different aspect ratios.
As illustrated in Fig.~\ref{fig: morevision}, ~\ref{fig: morevision2} and ~\ref{fig: morevision3}, our InpaintDPO generates high-quality inpainting results, with superior performance especially in background spatial rationality.

\begin{figure*}[t]
    \centering
    \includegraphics[width=0.9\linewidth]{figs/vision_result-12.pdf}
    \vspace{-0.1in}
    \caption{More cases generated by InpaintDPO. InpaintDPO supports the generation of high-quality images with different aspect ratios, while ensuring the reasonable placement of subjects, achieving realistic foregroundconditioned inpainting.}
    \label{fig: morevision2}
\end{figure*}

\begin{figure*}[t]
    \centering
    \includegraphics[width=0.9\linewidth]{figs/vision_result-13.pdf}
    \vspace{-0.1in}
    \caption{More cases generated by InpaintDPO. InpaintDPO supports the generation of high-quality images with different aspect ratios, while ensuring the reasonable placement of subjects, achieving realistic foregroundconditioned inpainting.}
    \label{fig: morevision3}
\end{figure*}

\begin{figure*}[t]
    \centering
    \includegraphics[width=0.85\linewidth]{figs/GPT4O-2.pdf}
    \vspace{-0.1in}
    \caption{GPT4o rationality analysis results}
    \label{fig: gpt4o}
\end{figure*}

\begin{figure*}[t]
    \centering
    \vspace{-0.1in}
    \includegraphics[width=0.77\linewidth]{figs/More_Vision_Result_1-6.pdf}
    \vspace{-0.1in}
    \caption{More qualitative comparisons between our InpaintDPO and the state-of-the-art methods. We highlight the unreasonable extension parts with \textcolor{Magenta}{pink} boxes, inappropriate positional relationship with \textcolor{green}{green} boxes and inappropriate viewpoint with \textcolor{orange}{orange} boxes, and label the inappropriate spatial scale with \textcolor{blue}{blue}. Please zoom in for more details.}
    \label{fig:more vision 1}
\end{figure*}

\begin{figure*}[t]
    \centering
    \vspace{-0.1in}
    \includegraphics[width=0.77\linewidth]{figs/More_Vision_Result_2-6.pdf}
    \vspace{-0.1in}
    \caption{More qualitative comparisons between our InpaintDPO and the state-of-the-art methods. We highlight the unreasonable extension parts with \textcolor{Magenta}{pink} boxes, inappropriate positional relationship with \textcolor{green}{green} boxes and inappropriate viewpoint with \textcolor{orange}{orange} boxes, and label the inappropriate spatial scale with \textcolor{blue}{blue}. Please zoom in for more details.}
    \label{fig:more vision 2}
\end{figure*}

\begin{figure*}[t]
    \centering
    \vspace{-0.1in}
    \includegraphics[width=0.77\linewidth]{figs/More_Vision_Result_3-6.pdf}
    \vspace{-0.1in}
    \caption{More qualitative comparisons between our InpaintDPO and the state-of-the-art methods. We highlight the unreasonable extension parts with \textcolor{Magenta}{pink} boxes, inappropriate positional relationship with \textcolor{green}{green} boxes and inappropriate viewpoint with \textcolor{orange}{orange} boxes, and label the inappropriate spatial scale with \textcolor{blue}{blue}. Please zoom in for more details.}
    \label{fig:more vision 3}
\end{figure*}

\begin{figure*}[t]
    \centering
    \includegraphics[width=0.85\linewidth]{figs/qwen_dpo_suppl-3.pdf}
    \vspace{-0.1in}
    \caption{More comparison between Qwen-Image-Edit and Qwen-InpaintDPO}
    \label{fig:qwen suppl}
\end{figure*}

% WARNING: do not forget to delete the supplementary pages from your submission 
% \input{sec/X_suppl}

% WARNING: do not forget to delete the supplementary pages from your submission 

\end{document}